\providecommand{\main}{./Results} 
\documentclass[%
  onecolumn
   , hidempi
]{mpi2015-cscpreprint}
\makeatletter
\DeclareOldFontCommand{\rm}{\normalfont\rmfamily}{\mathrm}
\DeclareOldFontCommand{\sf}{\normalfont\sffamily}{\mathsf}
\DeclareOldFontCommand{\tt}{\normalfont\ttfamily}{\mathtt}
\DeclareOldFontCommand{\bf}{\normalfont\bfseries}{\mathbf}
\DeclareOldFontCommand{\it}{\normalfont\itshape}{\mathit}
\DeclareOldFontCommand{\sl}{\normalfont\slshape}{\@nomath\sl}
\DeclareOldFontCommand{\sc}{\normalfont\scshape}{\@nomath\sc}
\makeatother

\usepackage[american]{babel}

\usepackage{graphicx}
\usepackage{amssymb}
\usepackage{amsthm}
\usepackage[mathscr]{eucal}

\numberwithin{equation}{section}
\numberwithin{figure}{section}
\numberwithin{table}{section}

\usepackage{mymacros}
\usepackage{import}
\usepackage{xifthen}
\usepackage{pdfpages}
\usepackage{transparent}
\usepackage{cleveref}
\usepackage{sidecap}    
\usepackage{floatrow}
\usepackage{todonotes}
\usepackage{algorithm}
\usepackage[noend]{algpseudocode}

\makeatletter
\def\algbackskip{\hskip-\ALG@thistlm}
\makeatother

\usepackage{caption}
\usepackage{subcaption}
\usepackage{nicematrix}
\usepackage{arydshln}

\definecolor{lightgray}{gray}{0.9}
\usetikzlibrary{fadings,shapes.arrows,shadows}   

\tikzfading[name=arrowfading, top color=transparent!0, bottom color=transparent!95]
\tikzset{arrowfill/.style={top color= black!20, bottom color=green!50!black, general shadow={fill=black, shadow yshift=-0.8ex, path fading=arrowfading}}}
\tikzset{arrowstyle/.style={draw=black,arrowfill, single arrow,minimum height=#1, single arrow,
		single arrow head extend=.2cm,}}

\newcommand{\tikzfancyarrow}[2][2cm]{\tikz[baseline=-0.0ex]\node [arrowstyle=#1] {#2};}

\newcommand{\rk}{{\sffamily{RK4}}}
\newcommand{\stdsindy}{{\sffamily{Std-SINDy}}}
\newcommand{\rksindy}{{\sffamily{RK4-SINDy}}}
\newcommand{\bred}[1]{{\color{red!50!black}#1}}
\newcommand{\dt}{\text{\sffamily{dt}}}

\begin{document}
  

\title{Discovery of Nonlinear Dynamical Systems using a Runge-Kutta Inspired Dictionary-based Sparse Regression Approach}
  
\author[$\ast$]{Pawan Goyal}
\affil[$\ast$]{Max Planck Institute for Dynamics of Complex Technical Systems, 39106 Magdeburg, Germany.\authorcr
  \email{goyalp@mpi-magdeburg.mpg.de}, \orcid{0000-0003-3072-7780}
}
  
\author[$\ast$]{Peter Benner}
\affil[$\dagger$]{Max Planck Institute for Dynamics of Complex Technical Systems, 39106 Magdeburg, Germany.\authorcr
  \email{benner@mpi-magdeburg.mpg.de}, \orcid{0000-0003-3362-4103}
}
  
\shorttitle{Discovery of Nonlinear Dynamical Systems using Sparse Regression}
\shortauthor{P. Goyal, P. Benner}
\shortdate{}
  

  
\abstract{%
	Discovering dynamical models to describe underlying dynamical behavior is essential to draw decisive conclusions and engineering studies, e.g., optimizing a process.  Experimental data availability notwithstanding has increased significantly, but interpretable and explainable models in science and engineering yet remain incomprehensible. In this work, we blend machine learning and dictionary-based learning with numerical analysis tools to discover governing differential equations from noisy and sparsely-sampled measurement data. We utilize the fact that given a dictionary containing huge candidate nonlinear functions, dynamical models can often be described by a few appropriately chosen candidates.  As a result, we obtain interpretable and parsimonious models which are prone to generalize better beyond the sampling regime. Additionally, we integrate a numerical integration framework with dictionary learning that yields differential equations without requiring or approximating derivative information at any stage. Hence, it is utterly effective in corrupted and sparsely-sampled data. We discuss its extension to governing equations, containing rational nonlinearities that typically appear in biological networks. Moreover, we generalized the method to governing equations that are subject to parameter variations and externally controlled inputs. We demonstrate the efficiency of the method to discover a number of diverse differential equations using noisy measurements, including a model describing neural dynamics, chaotic Lorenz model,  Michaelis-Menten Kinetics, and a parameterized Hopf normal form. 
}

\novelty{This work combines machine learning (dictionary-based) with a numerical integration scheme, namely a Runge-Kutta scheme to discover governing equations using corrupted and sparsely-sampled data. The method does not require the computation of derivative information to discover governing equations. Hence, it holds a key advantage when data are corrupted and sparsely sampled.} 
\maketitle

\section{Introduction}\label{sec:introduction}
Data-driven discovery of dynamic models has recently picked up much attention as there are revolutionary breakthroughs in data science and machine learning \cite{jordan2015machine,marx2013big}.  With the increasing ease of data availability and advances in machine learning, we can delve into analyzing data and identifying patterns to uncover dynamic models that faithfully describe the underlying dynamical behavior.  Though inferring dynamic models have been intensively studied in the literature, drawing conclusions and interpretations from them still remains strenuous.  Moreover, extrapolation and generalization of models are limited beyond the training regime. 

The sphere of identifying models using data is often referred to as system identification. For linear systems, there is an extensive collection of approaches \cite{lennart1999system,VanM96}.  However, despite several decades of research on learning nonlinear systems \cite{NarP90,SuyVdM96,kantz2004nonlinear}, it is still far away from being as mature as linear systems. Inferring nonlinear systems often require a prior model hypothesis by practitioners. A compelling breakthrough towards discovering nonlinear governing equations appeared in \cite{bongard2007automated,schmidt2009distilling}, where an approach based on genetic programming or symbolic regression is developed to identify nonlinear models using measurement data. It provides interpretable analytic models that accomplish a long-standing desire to the engineering community.  A parsimonious model is determined by examining the Pareto font that discloses a tread-off between the identified model's complexity and accuracy. In a similar spirit, there have been efforts to develop sparsity promoting approaches to discover nonlinear dynamical systems \cite{wang2011predicting,ozolicnvs2013compressed,proctor2014exploiting,brunton2016discovering,brunton2016sparse}. It is often observed that the dynamics of physical processes can be given by collecting a few nonlinear feature candidates from a high-dimensional nonlinear function space, referred to as a  feature dictionary. These sparsity-promoting methods are prone to discover models that are interpretable and parsimonious. 
Significant progress in solving sparse regression \cite{friedman2001elements,james2013introduction,tibshirani1996regression} and compressed sensing \cite{donoho2006compressed,candes2006robust,candes2006stable,tropp2007signal} support developments of these approaches. Although all these methods have gained much popularity, the success of these methods largely depends on the feature candidates included in the dictionary and the ability to accurately approximating the derivative information using measurement data. A derivative approximation using sparsely sampled and noisy measurements impose a tough challenge though there are approaches to deal with noise, see, e.g., \cite{chartrand2011numerical}
We also highlight additional directions explored in the literature to discover nonlinear governing equations, which include discovery of models using time-series data \cite{crutchfield1987equations}, automated inference of dynamics \cite{bongard2007automated,schmidt2011automated,daniels2015efficient}, and equation-free modeling \cite{kevrekidis2003equation,ye2015equation,proctor2014exploiting}. 

In this work, we re-conceptualize the problem of discovering nonlinear differential equations by blending sparse identification with a classical numerical integration tool. We here focus on a widely known integration scheme, namely  \emph{Runge-Kutta} $4^{\text{\sffamily th}}$-order \cite{ascher1998computer}. In contrast to previously studied sparse identification approaches, e.g., \cite{bongard2007automated,wang2011predicting,brunton2016discovering}, our approach would not require direct access or approximation of temporal gradient information. 
Therefore, we do not commit errors due to a gradient approximation. The approach becomes an attractive choice when the collected measurement data are sparsely sampled and corrupted with noise. We mention that numerical integration-inspired (e.g., Runge-Kutta) neural network architecture designs have also studied in the literature and have observed their supreme performances in deep learning, see, e.g., \cite{he2016deep,huang2017densely}, and from the perspective of dynamical modeling, see, e.g., \cite{raissi2018multistep,raissi2019physics,rudy2019deep}. These methods yield black-box models, thus interpretable and generalization of these models are ambiguous. 

What is more, we discuss an essential class of dynamic models that typically explains the dynamics of biological networks. It is also witnessed that regulatory and metabolic networks are sparse in nature, i.e., not all components influence each other. Furthermore, such dynamic models are often given by rational nonlinear functions. Consequently, the classical dictionary-based sparse identification ideology is not applicable as building all possible rational feature candidates is infeasible. To deal with this, the authors in \cite{mangan2016inferring} have recast the problem as finding the sparsest vector in a given null-space. However, computing a null space using corrupted measurement data is a non-trivial task though there is some work in the direction \cite{gavish2017optimal}.  In this work, we instead characterize identifying rational functions as a ratio of two functions, where each function is identified using dictionary learning. Hence, we inherently retain the primary principle of sparse identification in the course of discovering models.  In addition to these, we discuss the case where a dictionary contains parameterized candidates, e.g., $e^{\alpha x}$, where $x$ is dependent variables, and $\alpha$ is an unknown parameter. We extend our discussion to parametric and controlled dynamic processes.

The organization of the paper is as follows. In \Cref{sec:rksindy}, we briefly recap the  Runge-Kutta $4^\text{th}$-order scheme that is typically used to integrate differential equations. After that, we propose a methodology to discover differential equations by synthesizing the integration scheme with sparse identification.
Furthermore, since the method involves solving nonlinear and non-convex optimization problems that promote sparse solutions, \Cref{sec:algorithms} discusses algorithms inspired by a sparse-regression approach in \cite{tibshirani1996regression,brunton2016discovering}. 
In \Cref{sec:extension}, we examine a number of extensions to other classes of models, e.g., when governing equations are given by a ratio of two functions and involve model parameters and external control inputs. In the subsequent section, we illustrate the efficiency of the proposed methods by discovering a broad variety of benchmark examples, namely the chaotic Lorenz model, Fitz-Hugh Nagumo models, Michaelis-Menten Kinetics, and parameterized Hopf normal norm. We extensively study the performance of the proposed approach even under noisy measurements and compare it to the approach proposed in \cite{brunton2016discovering}. We conclude the paper with a summary and high-priority research directions.

\section{Discovering Nonlinear Governing Equations using a Runge-Kutta Inspired Sparse Identification} \label{sec:rksindy}
In this section, we are determined to discover nonlinear governing equations using sparsely sampled measurement data. These may be corrupted using experimental and/or sensor noise. We establish approaches by combining a numerical integration method and dictionary-based learning of the gradient field.
As a result, we develop methodologies that allow us to discover nonlinear differential equations without the explicit need for derivative information, unlike the approach proposed in \cite{brunton2016discovering,daniels2015efficient,wang2011predicting}. 
In this work, we utilize the widely employed approach to integrate differential equations, namely  \emph{Runge-Kutta $4^\text{th}$-order} (\rk) scheme, which is briefly outlined in the following. 
%
%
%


\subsection{Runge-Kutta $4^{\text{th}}$ order scheme}\label{subsec:rkdiscussion}
The \rk~scheme is a widely-used method to solve an initial value problem. Let us consider an initial value problem as follows: 
\begin{equation}\label{eq:NL_eq}
	\dot \bx(t) = \mathbf f (\bx(t)),\quad \bx(t_0) = \bx_0,
\end{equation}
where $\bx(t) := \begin{bmatrix} \bx_1(t),\bx_2(t),\ldots,\bx_n(t) \end{bmatrix}$ with $\bx_j(t)$ being the $j$th element of the vector $\bx(t)$. Assume that we aim at predicting $\bx(t_{k+1})$ for a given $\bx(t_k)$, where $k \in \{0,1,\ldots, \cN\}$. Then, using the \rk~scheme, $\bx(t_{k+1})$ can be given as a weighted sum of four increments, which are the product of the time-step and gradient field information $\mathbf f(\cdot)$ at the specific locations. Precisely, it is given as 
\begin{equation}\label{eq:RKstep}
	\bx(t_{k+1}) \approx \bx(t_k) + \dfrac{1}{6}h_k\left(\bk_1 + \bk_2 + \bk_3 + \bk_4\right), \quad h_k = t_{k+1} - t_k,
\end{equation}
where 
\begin{align*}
	\bk_1 &= \mathbf f(\bx(t_k) ),	\quad\bk_2 = \mathbf f\left(\bx\left(t_k + h_k\dfrac{\bk_1}{2}\right) \right), \quad
	\bk_3 = \mathbf f\left(\bx\left(t_k + h_k\dfrac{\bk_2}{2}\right) \right), \quad \bk_4 = \mathbf f\left(\bx\left(t_k + h_k\bk_3\right) \right).
\end{align*}
The \rk~scheme as a network is illustrated in \Cref{fig:rksteps}(a). 
The local integration error due to the \rk~scheme  is of $\cO(h_k^5)$; hence, the approach is very accurate for smaller time-steps. Furthermore,  if we integrate the equation \eqref{eq:NL_eq} from the time $t_0$ to $t_f$, we can  take $\cN$ steps with time-steps $h_k, k \in \{1,\ldots,\cN\}$ so that $t_f = t_0 + \sum_{i=0}^\cN h_k$.  In the rest of the paper, we use  a short-hand notation for the step  in \eqref{eq:RKstep} by $\cF_{\text{\rk}}\left(\mathbf f, \bx(t_k),h_k\right)$, i.e., 
\begin{equation}
	\bx(t_{k+1}) = \bx(t_{k} + h_k) \approx	\cF_{\text{\rk}}\left(\mathbf f, \bx(t_k),h_k\right).
\end{equation}
Lastly, we stress a point that the \rk~scheme readily handles integration backward in time, meaning that $h_k$ in \eqref{eq:RKstep} can also be negative. Hence, we can predict both $y(t_{k+1})$ and $y(t_{k-1})$ using $y(t_{k})$ very accurately using \rk~scheme. 
\subsection{Discovering nonlinear dynamical systems}
Next, we develop a \rk-inspired sparse identification approach to discover governing equations. Precisely, we aim at disclosing the most parsimonious representation of the gradient field $\mathbf f(\bx(t))$  in \eqref{eq:NL_eq} using only  a time-history of $\bx(t)$. Assume that the data is sampled at the time instances $\{t_0,\ldots,t_\cN\}$ and let us define time-steps $h_k:= t_{k+1} - t_{k}$. Furthermore, for simplicity of notation, we assume that the data follows \rk~exactly, but the method is not limited to it.  Consequently, we form two data matrices:
\begin{equation}\label{eq:constrctdataatrix}
	\begin{aligned}
		\bX&:=
		\begin{bmatrix}
			\bx(t_2) \\ \bx(t_3) \\ \vdots  \\ \bx(t_\cN) 
		\end{bmatrix} 
		= 
		\begin{bmatrix}
			\bx_1(t_2) & 	\bx_2(t_2) & \cdots & 	\bx_n(t_2) \\
			\bx_1(t_3) & 	\bx_2(t_3) & \cdots & 	\bx_n(t_3) \\
			\vdots & \vdots & \ddots & \vdots \\
			\bx_1(t_\cN) & 	\bx_2(t_\cN) & \cdots & 	\bx_n(t_\cN) \\
		\end{bmatrix}& \text{and}~~
		\bX_\cF(\mathbf f) &:=
		\begin{bmatrix}
			\cF_{\text{\rk}}\left( \mathbf f, \bx(t_0), h_1\right) \\ \cF_{\text{\rk}}\left( \mathbf f, \bx(t_1), h_2\right) \\ \vdots  \\ \cF_{\text{\rk}}\left( \mathbf f, \bx(t_{\cN-1}), h_{\cN}\right)
		\end{bmatrix}.
	\end{aligned}
\end{equation}
The next important ingredients to sparse identification is the construction of  a huge symbolic dictionary $\mathbf \Phi$, containing potential nonlinear features. So, the function $\mathbf f(\cdot)$ can be given by a linear combination of few terms from the dictionary. For example, one can consider a dictionary containing, polynomial, exponential, and trigonometric functions, which, for any given vector $\bv := \begin{bmatrix} \bv_1,\ldots, \bv_n \end{bmatrix} $ can be given as:
\begin{equation}
	\mathbf\Phi(\bv) = \begin{bmatrix}  1, \bv, \left.\bv^{\mathscr{P}_2}\right., \left.\bv^{\mathscr P_3}\right.,\ldots, \mathbf e^{-\bv}, \mathbf e^{-2\bv}, \ldots, \boldsymbol \sin(\bv), \boldsymbol\cos(\bv),\ldots \end{bmatrix}
\end{equation}
in which $\bv^{\mathscr P_i}, i \in \{2,3\}$ denote high-order polynomials, e.g., $\bv^{\mathscr P_2}$ contains all possible degree-2 polynomials of elements of $\bv$ as:
\begin{equation}\label{eq:degree_two}
	\bv^{\mathscr P_2} = \begin{bmatrix}  \bv_1^2, \bv_1\bv_2,\ldots, \bv_2^2, \bv_2\bv_3,\ldots, \bv_n^2	\end{bmatrix}
\end{equation}
Each element in the dictionary $\boldsymbol\Phi$ is a potential candidate to describe the function $\mathbf f$. Moreover, depending on applications, one may take the help of experts and include empirical knowledge to construct a meaningful feature dictionary. 

Having paradise of an extensive dictionary, one has many choices to choose candidates from the dictionary. However, our goal is to choose as few candidates as possible, describing the nonlinear function $\mathbf f$ in \eqref{eq:NL_eq}. Hence, we set up a sparsity-promoting optimization problem to pick few candidate functions from the dictionary, e.g., 
\begin{equation}
	\begin{aligned}
		\mathbf f_k(\bx(t)) &=  \boldsymbol \Phi(\bx(t)) \boldsymbol{\theta}_k,
	\end{aligned}
\end{equation}
where $\mathbf f_k: \Rn \rightarrow \R$ is the $k$th element of $\mathbf f$, and $\boldsymbol\theta_k$ a sparse vector; hence, selecting appropriate candidates from the dictionary determines governing equations. As a result, we can write the function $\mathbf f(\cdot)$ in \eqref{eq:NL_eq} as follows:
\begin{equation}
	\mathbf f(\bx) = \begin{bmatrix}
		\mathbf f_1(\bx),~ \mathbf f_2(\bx),~ \ldots,~\mathbf f_n(\bx) 
	\end{bmatrix} = 
	\begin{bmatrix}
		\boldsymbol \Phi(\bx) \boldsymbol{\theta}_1, \boldsymbol \Phi(\bx) \boldsymbol{\theta}_2, \cdots, \boldsymbol \Phi(\bx) \boldsymbol{\theta}_n
	\end{bmatrix} = 
	\boldsymbol \Phi(\bx)\boldsymbol{\Theta},
\end{equation}
where $\boldsymbol{\Theta} = \begin{bmatrix}  \boldsymbol{\theta}_1^\top, \ldots, \boldsymbol{\theta}_n  \end{bmatrix}$. 
\begin{figure}[tb]
	\centering
	\begin{tikzpicture}[font=\sffamily]
		\node[ fill = white!95!black,draw = green!50!black, text = white, thick,rounded corners = 0.5ex] (rk4) {
			\includegraphics[height = 4.0cm]{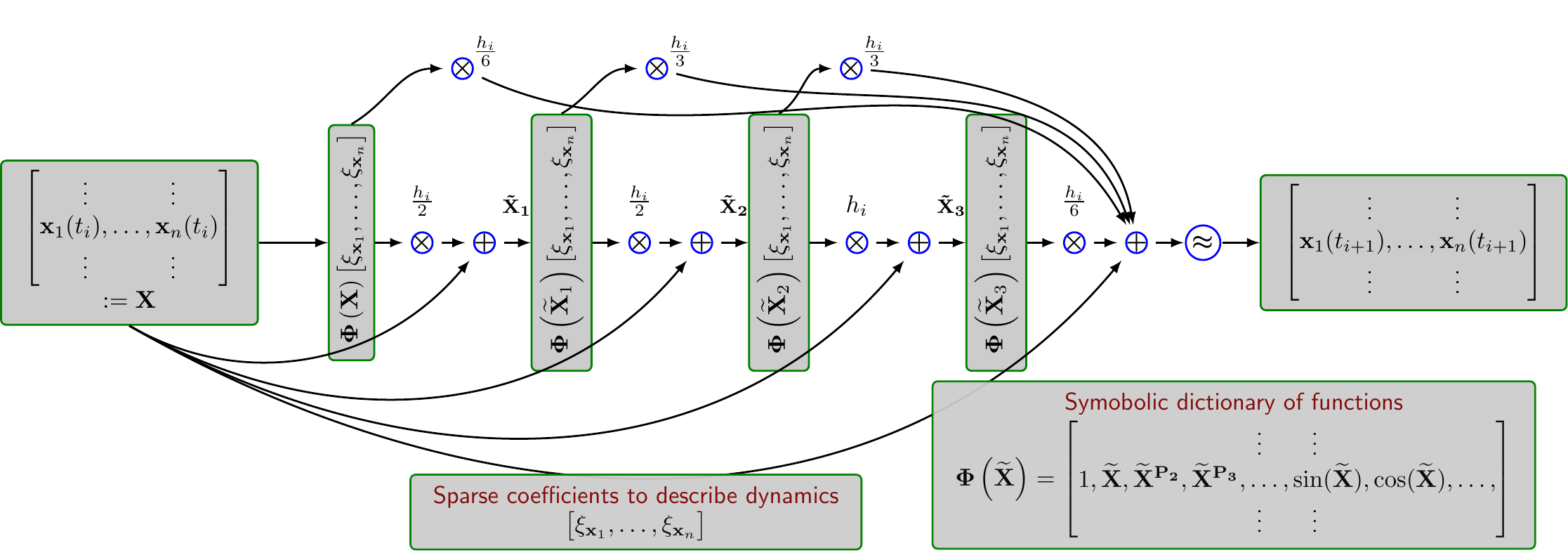}
		};
		\node[ fill = white!95!black,draw = green!50!black, text = white, thick,below = 0.1cm of rk4,rounded corners = 0.5ex] (FHN) {
			\includegraphics[height = 4.0cm]{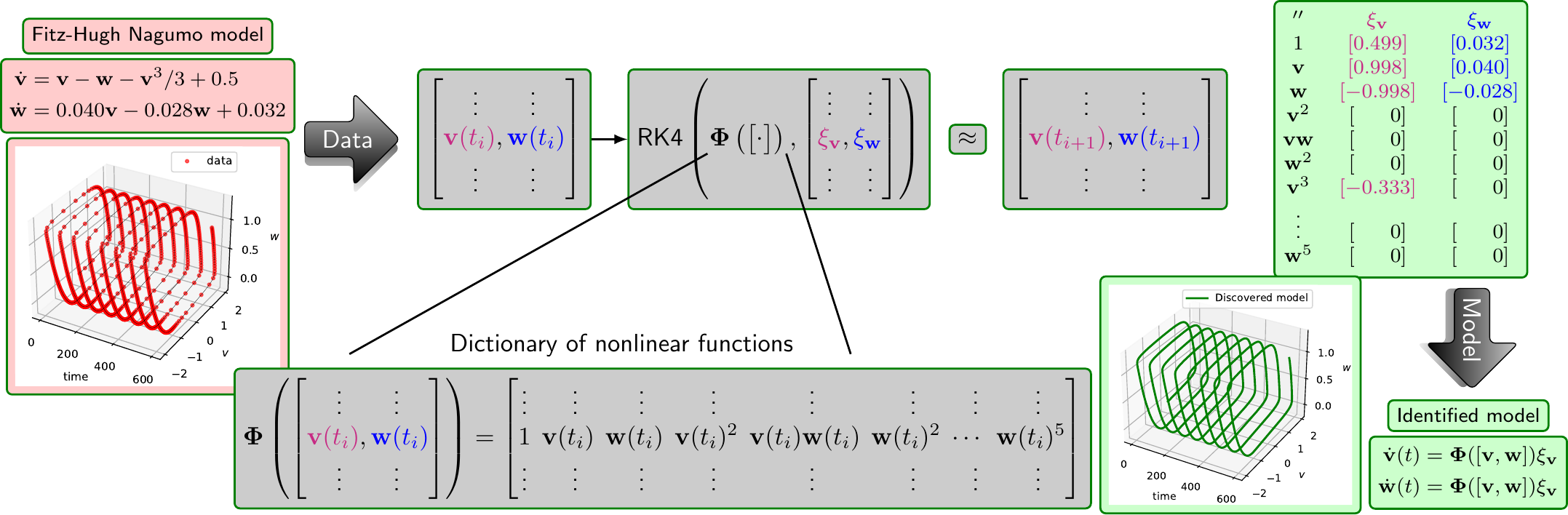}
		};
		\node[ fill = white!75!black,draw = green!50!black, text = green!50!black, thick, below left = -.7cm and -1cm of rk4,rounded corners = 0.5ex] (a) {
			(a)};
		\node[ fill = white!75!black,draw = green!50!black, text = green!50!black, thick, below left = -.7cm and -1cm of FHN,rounded corners = 0.5ex] (b) {
			(b)};
	\end{tikzpicture}
	\caption{In (a), we show the \rk~scheme to predict variables at the next time-step as a network. It resembles a residual-type network with skip connections. In (b), we present a systematic illustration of \rksindy~approach to discover governing equations  using the Fitz-Hugh Nagumo model. In the first step, we collect a time history of variables $\bv(t)$ and $\bw(t)$. Next, we build a symbolic feature dictionary $\boldsymbol{\Phi}$, containing potential features. It is followed by solving a nonlinear sparse regression problem to pick the right features from the dictionary (encoded in sparse vectors $\xi_\bv$ and $\xi_\bw$). Here, we presume that variables at the next time steps are given by following the \rk~scheme. 
		The non-zero entries in vectors $\xi_\bv$ and $\xi_\bw$ determine the right-hand side of the differential equations. As shown, we pick the right features from the dictionary upon solving the optimization problem, and corresponding coefficients are $0.1\%$ accurate.} 
	\label{fig:rksteps}
\end{figure}
This allows to articulate our optimization problem that aims at discovering governing equations -- that is to find the sparsest $\boldsymbol{\Theta}$, satisfying
\begin{equation}\label{eq:opt_problem}
	\bX = \bX_\cF\left(\mathbf f\right), ~~\text{where}~~\mathbf f(\bx) = \boldsymbol \Phi(\bx)\boldsymbol{\Theta}.
\end{equation}
Once we identify $\boldsymbol{\Theta}$  or $\{\boldsymbol\theta_1,\ldots,\boldsymbol\theta_n\}$, the dynamic model can be given as
\begin{equation*}
	\begin{bmatrix}
		\bx_1(t),~ \bx_2(t),~  \ldots,~ \bx_n(t)
	\end{bmatrix} = 
	\begin{bmatrix}
		\boldsymbol \Phi(\bx)\boldsymbol\theta_1,~\boldsymbol \Phi(\bx)\boldsymbol\theta_2,~  \ldots, \boldsymbol \Phi(\bx)\boldsymbol\theta_n
	\end{bmatrix}.
\end{equation*}
We referred to the proposed approach as Runge-Kutta inspired sparse identification (\rksindy). 
We depict all the essential steps for \rksindy~to discover governing equations in \Cref{fig:rksteps} through the Fitz-Hugh Nagumo model (details of the model are provided later).  

We take the opportunity to stress imperative advantages of \rksindy. That is -- to discover nonlinear differential equations, we do not require derivative information of $\bx(t)$ at any step. We only hypothesize that the gradient field can be given by selecting appropriate features from a dictionary containing a vast number of possible nonlinear features. Consequently, we expect to discover good quality models when data are sparsely collected and/or are corrupted, and this is what we manifest in our results in \Cref{sec:results}.  Interestingly, the approach readily handles irregular time-steps. 

When the data are corrupted with noise or does not follow \rk~exactly, then we may need to regularize the above optimization problem. Since the $l_1$-regularization  promotes sparsity in the solution, one can solve an $l_1$-regularized  optimization problem:
\begin{equation}\label{eq:opt_problem1}
	\min_{\boldsymbol{\Theta}} 	\|\bX - \bX_\cF\left(\boldsymbol \Phi(\cdot)^\otimes\boldsymbol{\Theta}\right)\| + \lambda\left\| \boldsymbol{\Theta}\right\|_{l_1}.
\end{equation}

As discussed in Subsection \ref{subsec:rkdiscussion}, the \rk~scheme can accurately predict both $\bx(t_{i+1})$ and $\bx(t_{i-1})$ using $\bx(t_{i})$. Therefore, the following also holds:
\begin{equation*}
	\bX^{\text{\sffamily{b}}} = \bX_\cF^{\text{\sffamily{b}}}\left(\mathbf f\right),
\end{equation*}
where
\begin{equation*}
	\begin{aligned}
		\bX^{\text{\sffamily{b}}}&:=
		\begin{bmatrix}
			\bx(t_0) \\ \bx(t_1) \\ \vdots  \\ \bx(t_{\cN-1}) 
		\end{bmatrix} 
		= 
		\begin{bmatrix}
			\bx_1(t_0) & 	\bx_2(t_0) & \cdots & 	\bx_n(t_0) \\
			\bx_1(t_1) & 	\bx_2(t_1) & \cdots & 	\bx_n(t_2) \\
			\vdots & \vdots & \ddots & \vdots \\
			\bx_1(t_{\cN-1}) & 	\bx_2(t_{\cN-1}) & \cdots & 	\bx_n(t_{\cN-1}) \\
		\end{bmatrix}& \text{and}~~
		\bX_\cF^{\text{\sffamily{b}}}(\mathbf f) &:=
		\begin{bmatrix}
			\cF_{\text{\rk}}\left( \mathbf f, \bx(t_1), -h_1\right) \\ \cF_{\text{\rk}}\left( \mathbf f, \bx(t_2), -h_2\right) \\ \vdots  \\ \cF_{\text{\rk}}\left( \mathbf f, \bx(t_{\cN}), -h_{\cN}\right)
		\end{bmatrix}.
	\end{aligned}
\end{equation*}
Therefore, we can have a more involved optimization by including both forward and backward predictions in time. This helps particularly in noisy measurement data. 
In the next subsection, we discuss an efficient procedure to solve the optimization problem \eqref{eq:opt_problem}.

\section{Algorithms to Solve Nonlinear  Sparse Regression Problems}\label{sec:algorithms}
Several methodologies exist to solve linear optimization problems that yield a sparse solution, see ,e.g., LASSO \cite{tibshirani1996regression,friedman2001elements}. However, the optimization problem \eqref{eq:opt_problem} is nonlinear and likely non-convex. There are some developments in solving sparsity-constrained nonlinear optimization problems; see, e.g., \cite{beck2013sparsity,yang2016sparse}. Though these methods enjoy many nice theoretical properties, they typically require a priory the maximum number of non-zero elements in the solutions, which is often unknown to us. Also, they are computationally demanding.  

Here, we propose two simple gradient-based sequential thresholding schemes, similar to the one discussed in \cite{brunton2016discovering} for linear problems.  In these schemes, we first solve the nonlinear optimization problem \eqref{eq:opt_problem} using a (stochastic-) gradient descent method to obtain $\boldsymbol{\Theta}_1$, followed by applying a thresholding to~$\boldsymbol{\Theta}_1$.

\subsection{Fix cutoff thresholding} In the first approach, we define a cutoff value $\lambda$ and set all the coefficients smaller than $\lambda$ to zero. We then update the remaining non-zero coefficients by solving the optimization problem \eqref{eq:opt_problem} again, followed by employing the thresholding. We repeat the procedure until all the non-zero coefficients are equal to or larger than $\lambda$. This procedure is efficient as the current value of non-zero coefficients can be used as an initial guess for the next iteration, and the optimal $\boldsymbol{\Theta}$ can be found with a little computational effort. Note the cutoff parameter $\lambda$ is important to obtain a suited sparse solution, but it can be found using the concept of cross-validation. We sketch the discussed procedure in \Cref{algo:procedure1}. 

\begin{algorithm}[tb]
	\caption{Fix Cutoff Thresholding Procedure}
	\begin{flushleft}
		\textbf{Input:} Measurement data $\{\bx(t_0), \bx(t_1),\ldots, \bx(t_\cN)\}$ and the cutoff parameter $\lambda$. \\[-100pt]
	\end{flushleft}\vspace{-.25cm}
	\begin{algorithmic}[1]
		\State Solve the optimization problem \eqref{eq:opt_problem} to get $\boldsymbol{\Theta}$.
		\State \texttt{small\_idx} = $\left(\abs{\boldsymbol\Theta} < \lambda\right)$ \Comment{Determine indices at which coefficients are less $\lambda$}
		\State \texttt{Err} = $\|\boldsymbol{\Theta}\left(\texttt{small\_idx}\right)\|$
		\While{$\texttt{Err} \neq 0$}
		\State Update $\boldsymbol{\Theta}$ by solving the optimization problem \eqref{eq:opt_problem} with the constraint $\boldsymbol{\Theta}\left(\texttt{small\_idx}\right)=0$		
		\State \texttt{small\_idx} = $\left(\abs{\boldsymbol\Theta} < \lambda\right)$ \Comment{Determine indices at which coefficients are less $\lambda$}
		\State \texttt{Err} = $\|\boldsymbol{\Theta}\left(\texttt{small\_idx}\right)\|$
		\EndWhile  \label{algo:procedure1}
	\end{algorithmic}
	\begin{flushleft}\vspace{-.25cm}
		\textbf{Output:} The sparse $\boldsymbol{\Theta}$ that picks right features from the dictionary.
	\end{flushleft}\vspace{-.25cm}
\end{algorithm}

\subsection{Iterative cutoff thresholding} In the fix cutoff thresholding approach, we need to pre-define the cutoff value for thresholding. A suitable value of it needs to be found by an iterative procedure. In our empirical observations, applying fix thresholding at each iteration does not yield the most sparse solution in many instances. To circumvent this, we propose an iterative way of thresholding -- that is as follows. In the first step, we solve the optimization problem \eqref{eq:opt_problem} for $\boldsymbol{\Theta}$. Then, we determine the smallest non-zero coefficients of $\abs{\boldsymbol\Theta}$ followed by setting all the coefficients smaller than this to zero. Like the previous approach, we update the remaining non-zero coefficients by solving the optimization problem \eqref{eq:opt_problem}. We repeat the step of finding the smallest non-zero coefficient of the updated $\abs{\boldsymbol{\Theta}}$ and setting it to zero. We iterate the procedure until the loss of data fidelity is less than a given tolerance.  Visually, it can be anticipated using the curve between the data-fitting and number of non-zero elements in $\Theta$, which typically exhibit an \emph{elbow}-type curve. We shall see in our result section (\Cref{sec:results}).  We sketch the step of the procedure in \Cref{algo:procedure2}. 

We note that the successive iterations converge faster to the optimal value after the first thresholding as we choose the coefficients after applying thresholding as the initial guess. Moreover, in our experiments, we observe that this thresholding approach yields better results, particularly when data are corrupted with noise. However, it may be computationally more expensive than the fixed cutoff thresholding approach as it may need more iterations to converge. Therefore, an efficient approach combining fixed and iterative thresholding approaches is a worthy future research direction. 

\begin{algorithm}[tb]
	\caption{Iterative Cutoff Thresholding Procedure}
	\begin{flushleft}
		\textbf{Input:} Measurement data $\{\bx(t_0), \bx(t_1),\ldots, \bx(t_\cN)\}$. \\[-100pt]
	\end{flushleft}\vspace{-.25cm}
	\begin{algorithmic}[1]
		\State Construct $\bX$ using measurement data as in \eqref{eq:constrctdataatrix}.
		\State Solve the optimization problem \eqref{eq:opt_problem} to get $\boldsymbol{\Theta}$.
		\State $\cE := \|\bX - \bX_\cF\left(\boldsymbol \Phi(\cdot)\boldsymbol{\Theta}\right)\| $, where $\bX_\cF$ is defined in \eqref{eq:constrctdataatrix}.
		\While{$\cE \leq \texttt{tol}$} 
		\State Determine the smallest non-zero coefficient of $\texttt{abs}(\boldsymbol{\Theta})$, denoted by $\lambda_{\texttt{small}}$.
		\State \texttt{small\_idx} = $\left(\abs{\boldsymbol\Theta} <  \lambda_{\texttt{small}}\right)$ \Comment{Determine indices at which coefficients are less $\lambda$}
		\State Update $\boldsymbol{\Theta}$ by solving the optimization problem \eqref{eq:opt_problem} with the constraint $\boldsymbol{\Theta}\left(\texttt{small\_idx}\right)$.
		\State $\cE := \|\bX - \bX_\cF\left(\boldsymbol \Phi(\cdot)\boldsymbol{\Theta}\right)\| $.
		\EndWhile
		\label{algo:procedure2}
	\end{algorithmic}
	\begin{flushleft}\vspace{-.25cm}
		\textbf{Output:} The sparse $\boldsymbol{\Theta}$ that picks right features from the dictionary.
	\end{flushleft}\vspace{-.25cm}
\end{algorithm}

\section{A Number of Possible Extensions}\label{sec:extension}
In this section, we discuss several extensions to the methodology proposed in \Cref{sec:rksindy},  generalizing to a large class of problems. First, we discuss the discovery of governing differential equations given by a ratio of two functions. 
Next, we investigate the case in which a symbolic dictionary is parameterized. This is of particular interest when governing equations are expected to have candidate features, e.g., $\mathbf e^{\alpha \bx(t)}$, where $\alpha$ is unknown. 
We further extend our discussion to parameterized and externally controlled governing equations.
\subsection{Governing equations as a ratio of  two functions}\label{sec:rational}
There are many instances, where the governing equations  are given as a ratio of two nonlinear functions. Such equations frequently appear in the modeling of biological networks. For simplicity, we here examine a scalar problem; however, the extension to multi-dimensional cases readily follows.  Consider governing equations of the form:
\begin{equation}\label{eq:ratio_fun}
	\dot\bx(t) = \dfrac{\bg(\bx)}{1+\bh(\bx)},
\end{equation}
where $\bg(\bx) : \R \rightarrow \R$ and $\bh(\bx) : \R \rightarrow \R$ are continuous nonlinear functions.  Here again, the observation is that the functions $\bg(\cdot)$ and $\bh(\cdot)$ can be given as linear combinations of a few terms from corresponding dictionaries. Hence, we can cast the problem of identifying the model \eqref{eq:ratio_fun} as a dictionary-based discovery of governing equations. Let us consider two symbolic dictionaries:
\begin{align}
	\boldsymbol{\Phi}^{(\bg)}(\bx) & = \begin{bmatrix} 1, \bx, \bx^2,\bx^3, \ldots, \sin{(\bx)}, \cos{(\bx)}, \sin{(\bx^2)}, \cos{(\bx^2)}, \sin{(\bx^3)}, \sin{(\bx^3)},\ldots \end{bmatrix},\\
	\boldsymbol{\Phi}^{(\bh)}(\bx) & = \begin{bmatrix} \bx, \bx^2,\bx^3, \ldots, \sin{(\bx)}, \cos{(\bx)}, \sin{(\bx^2)}, \cos{(\bx^2)}, \sin{(\bx^3)}, \sin{(\bx^3)},\ldots \end{bmatrix}.
\end{align}
Consequently, the functions $\bg(\cdot)$ and $\bh(\cdot)$ can be given by
\begin{align}
	\bg(\bx) &= 	\boldsymbol{\Phi}^{(\bg)}(\bx)\boldsymbol{\theta}_\bg,\\
	\bh(\bx) &= 	\boldsymbol{\Phi}^{(\bh)}(\bx)\boldsymbol{\theta}_\bh,
\end{align}
where $\theta_\bg$ and $\theta_\bh$ are sparse vectors. 
Then, we can readily apply the framework discussed in the previous section by assuming $\mathbf f(\bx) := \dfrac{\bg(\bx)}{1+\bh(\bx)}$ in \eqref{eq:NL_eq}. We can determine sparse coefficients $\boldsymbol{\theta}_\bg$ and $\boldsymbol{\theta}_\bh$ by employing the thresholding  concepts  presented in \Cref{algo:procedure1,algo:procedure2}.  These are  possible because the algorithms are gradient-based and we only need to compute gradients with respect to $\boldsymbol{\theta}_\bg$ and $\boldsymbol{\theta}_\bh$.

Furthermore, we notice that it is worthwhile to consider governing equations of the form:
\begin{equation}\label{eq:gen_form}
	\dot{\bx}(t) =  \bk(\bx) + \dfrac{\bg(\bx)}{1+ \bh(\bx)}. 
\end{equation}
Indeed, the model \eqref{eq:gen_form} can be rewritten in the form considered in \eqref{eq:ratio_fun}. But it is rather efficient to consider the form \eqref{eq:gen_form}.  We illustrate it with the following example:
\begin{equation}\label{eq:demo_rat}
	\dot	\bx(t) = -\bx(t) -  \dfrac{\bx(t)}{1+\bx(t)},
\end{equation}
which fits to the form considered in \eqref{eq:gen_form}. In this case, all nonlinear functions $\bk(\cdot), \bg(\cdot)$ and $\bh(\cdot)$ are of degree-1 polynomials. On the other hand, if the model \eqref{eq:demo_rat} is written in the form \eqref{eq:ratio_fun}, then we have 
\begin{equation}
	\dot \bx(t) =    \dfrac{-1 -\bx(t) - \bx(t)^2}{1+\bx(t)}.
\end{equation}
Thus, the nonlinear functions $\bg(\cdot)$ and $\bh(\cdot)$ in \eqref{eq:ratio_fun} are of degrees 2 and 1, respectively. This gives a hint that if we aim at learning governing equations using sparse identification, it might be efficient to consider the form \eqref{eq:gen_form} from the complexity of the dictionary. It becomes even more adequate in multi-dimensional differential equations.  To discover dynamic model of the form \eqref{eq:gen_form}, we extend the idea of learning nonlinear functions using dictionaries. Let us construct three dictionaries as follows:

\begin{align}
	\boldsymbol{\Phi}^{(\bk)}(\bx) & = \begin{bmatrix} 1, \bx, \bx^2,\bx^3, \ldots, \sin{(\bx)}, \cos{(\bx)}, \sin{(\bx^2)}, \cos{(\bx^2)}, \sin{(\bx^3)}, \sin{(\bx^3)},\ldots \end{bmatrix},\\
	\boldsymbol{\Phi}^{(\bg)}(\bx) & = \begin{bmatrix} 1, \bx, \bx^2,\bx^3, \cdots, \sin{(\bx)}, \cos{(\bx)}, \sin{(\bx^2)}, \cos{(\bx^2)}, \sin{(\bx^3)}, \sin{(\bx^3)},\cdots \end{bmatrix},\\
	\boldsymbol{\Phi}^{(\bh)}(\bx) & = \begin{bmatrix} \bx, \bx^2,\bx^3, \ldots, \sin{(\bx)}, \cos{(\bx)}, \sin{(\bx^2)}, \cos{(\bx^2)}, \sin{(\bx^3)}, \sin{(\bx^3)},\cdots \end{bmatrix}.
\end{align}

Then, we believe that the nonlinear functions in \eqref{eq:gen_form} can be given as a sparse linear combination of the dictionaries, i.e., 
\begin{align}
	\bk(\bx) &= 	\boldsymbol{\Phi}^{(\bk)}(\bx)\boldsymbol{\theta}_\bk,&\quad
	\bg(\bx) &= 	\boldsymbol{\Phi}^{(\bg)}(\bx)\boldsymbol{\theta}_\bg,&\quad
	\bh(\bx) &= 	\boldsymbol{\Phi}^{(\bh)}(\bx)\boldsymbol{\theta}_\bh.
\end{align}
To determine the sparse coefficients $\{\boldsymbol{\theta}_\bk,\boldsymbol{\theta}_\bg,\boldsymbol{\theta}_\bh\}$, we can employ the \rksindy~framework, and \Cref{algo:procedure1,algo:procedure2}. We will illustrate this approach to discover an enzyme kinetics in \Cref{subsec:MM_model} that is given as a rational function.

\subsection{Discovering of parametric and externally controlled equations}
The \rksindy~immediately embraces the discovering of governing equations that are parametric and externally controlled. Let us begin with an externally controlled dynamic models of the form:
\begin{equation}\label{eq:controlledmodels}
	\dot \bx(t) = \mathbf f(\bx(t),\bu(t)),
\end{equation}
where $\bx(t) \in \Rn$ and $\bu(t)\in \Rm$ are state and controlled input vectors.  The goal here is to discover $\mathbf f(\bx(t),\bu(t))$ using the state trajectory $\bx(t)$ generated using a controlled input $u(t)$.  We aim at discovering governing equations using dictionary-based identification. Like discussed in \Cref{sec:rksindy}, we construct a  symbolic dictionary $\boldsymbol{\Phi}$ of possible candidate features using $\bx$ and $\bu$, i.e., 
\begin{equation}
	\boldsymbol{\Phi}(\bx,\bu) =  \begin{bmatrix} 1,\bx^\top, \bu^\top, \left(\bx_\bu^{\mathscr{P}_2}\right)^\top, \left(\bx_\bu^{\mathscr P_3}\right)^\top \end{bmatrix},
\end{equation}
where $\left(\bx_\bu^{\mathscr P_i}\right)^\top$ consists polynomial terms of degree-$i$, i.e., $\left(\bx_\bu^{\mathscr P_2}\right)^\top$ contains  degree-$2$ polynomial terms including cross terms:
\begin{equation}
	\left(\bx_\bu^{\mathscr P_2}\right)^\top = \begin{bmatrix}  \bx_1^2, \ldots, \bx_n^2,\bu_1^2,  \ldots, \bu_m^2,  \bx_1\bu_1, \ldots, \bx_n\bu_1, \bx_1\bu_2,\ldots, \bx_n\bu_m \end{bmatrix},
\end{equation}
where $\bu_i$ is the $i$-th element of $\bu$.
Using measurements of $\bx$ and $\bu$, we can cast the problem exactly as done in \Cref{sec:rksindy} by assuming that $\mathbf f(\bx(t),\bu(t))$ can be determined by selecting appropriate functions from the dictionary $\boldsymbol{\Phi}(\bx,\bu)$. Similarly, system parameters can also be incorporated to discover parametric differential equations of the form:
\begin{equation}
	\dot\bx(t) = \mathbf f(\bx(t),\mu),
\end{equation}
where $\mu\in \Rp$ is the system parameters. It can be considered as a special case of \eqref{eq:controlledmodels} since a constant input can be thought of as a parameter in the course of discovering governing equations. We illustrate \rksindy~for discovering parametrized Hopf normal form using measurement data (see Subsection \ref{subsec:hopf}).

\subsection{Parameterized dictionary}\label{sec:variabledictionary}
The success of the sparse identification highly depends on the quality of a constructed feature dictionary. In other words, the dictionary should contain right features in which governing differential equations can be given as a linear combination of few terms from the dictionary. However, it becomes a challenging task when one aims at including, for instance, trigonometry or exponential functions (e.g., $\sin{(ax)}, \mathbf e^{(bx)}$), where $\{a,b\}$ are unknown. In an extreme case, one might think of including $\sin(\cdot)$ and $\mathbf e{(\cdot)}$ for each possible value of $a$ and $b$. This would lead to the dictionary of infinite dimension, hence becomes intractable. To illustrate it, we consider the governing equation as follows:
\begin{equation}\label{eq:demo_exp}
	\dot \bx(t) = -\bx(t) + \exp(-1.75\bx(t)).
\end{equation}
Let us assume that we concern about discovering the model \eqref{eq:demo_exp} using a time history of $\bx(t)$ without any prior knowledge except that we expect exponential nonlinearities. It may be gathered with the help of experts or from empirical knowledge. For instance, an electrical circuit modeling containing diode components typically involves exponential nonlinearities, but the corresponding coefficient is unknown. 

We conventionally build a dictionary containing exponential functions using several possible coefficients as follows:
\begin{equation}
	\boldsymbol{\Phi}(\bx) = \begin{bmatrix}
		1, \bx, \bx^2, \bx^3, \ldots, \mathbf e^{\bx}, \mathbf e^{-\bx},\mathbf e^{2\bx}, \mathbf e^{-2\bx}\ldots, \sin(\bx), \cos(\bx), \ldots
	\end{bmatrix}.
\end{equation} 
However, it is impossible to add all infinitely exponential terms with different coefficients in the dictionary. As a remedy, we discuss the idea of a parameterized dictionary that was also discussed in \cite{champion2020unified}:
\begin{equation}
	\boldsymbol{\Phi}_\Xi(\bx) = \begin{bmatrix}
		1, \bx, \bx^2, \bx^3, \ldots, \sin(\eta_1\bx), \cos(\eta_2\bx),\sin(\eta_3\bx^2), \cos(\eta_4\bx^2), \ldots, \mathbf e^{\eta_5\bx}, \mathbf e^{\eta_6\bx^2}, \ldots,
	\end{bmatrix}
\end{equation} 
where $\Xi = \{\xi_1,\xi_2,\ldots\}$. In this case, we do not need to include all frequencies for trigonometric functions and coefficients for exponential functions. However, it comes at the cost of finding suitable coefficients $\{\eta_i\}$'s, along with a vector, selecting right features from the dictionary. Since we solve optimization problems, e.g., \eqref{eq:opt_problem} using a gradient descent, we can easily incorporate the parameters $\eta_i$'s along with $\theta_i$'s as learning parameters and can readily employ  \Cref{algo:procedure1,algo:procedure2} with a little alteration.  

\section{Results}\label{sec:results}
Here, we demonstrate the success of  \rksindy~to discover governing equations using measurement data through a number of examples of different complexity\footnote{Most of all examples are taken from \cite{brunton2016discovering}}. In the first example, we consider simple illustrative examples, namely, linear and nonlinear damped oscillators. Using the linear damped oscillator, we perform a comprehensive study under various conditions, i.e., the robustness of the approach to sparsely sampled and highly corrupted data. We compare the performance of our approach to discover governing equations with  \cite{brunton2016discovering}; we refer to it as \stdsindy\footnote{We use the Python implementation of the method, the so-called \texttt{PySINDy} \cite{desilva2020}.}. In the second example, we study the chaotic Lorenz example and show that \rksindy~determines the governing equations, exhibiting the chaotic behavior accurately. In the third example, we discover neural dynamics from measurement data using \rksindy. As the fourth example, we illustrate the discovery of a model that describes the dynamics of enzyme activity and contains rational nonlinearities. In the last example, we showcase that \rksindy~also successfully discovers the parametric Hopf normal form from collected noisy measurement data for various parameters.  


\subsection{Two-dimensional Damped Oscillators}
As simple illustrative examples, we consider two-dimensional damped harmonic oscillators. These can be given by linear and nonlinear models. We begin by considering the linear one.
\subsubsection{Linear damped oscillator}
Consider a 2D linear damped oscillator whose dynamics is given by:
\begin{subequations}
	\begin{align}
		\dot \bx(t)  & = -0.1 \bx(t) + 2.0\by(t),\\
		\dot \by(t) &= -2.0\bx(t) -0.1\by(t).
	\end{align}
\end{subequations}
To infer governing equations from measurement data, we first assume to have clean data at a regular time-step $\dt$. We then build a symbolic dictionary containing polynomial nonlinearities up to the degree of $5$. Next, we learn governing equations using \rksindy~(\Cref{algo:procedure1} with $\lambda = 5\cdot 10^{-2}$) and observe the quality of inferred equations for different $\dt$. We also present a comparison with \stdsindy. 

\begin{figure}[H]
	\centering
	\begin{subfigure}[b]{0.495\textwidth}
		\centering
		\includegraphics[width=1\textwidth]{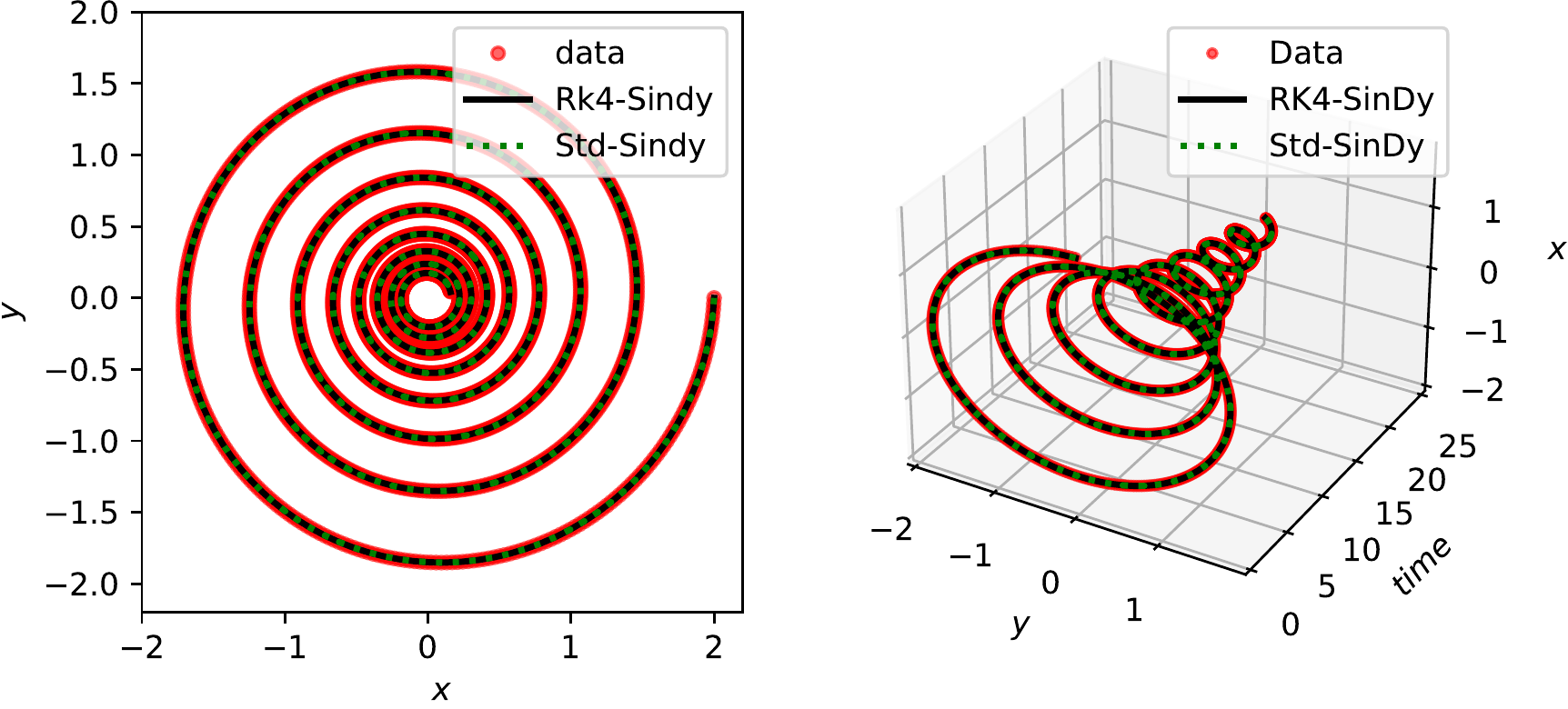}
		\caption{Time step $\dt = 1\cdot 10^{-2}$.}
		\label{fig:linear2d_dt_01}
	\end{subfigure}
	\begin{subfigure}[b]{0.495\textwidth}
		\centering
		\includegraphics[width=1\textwidth]{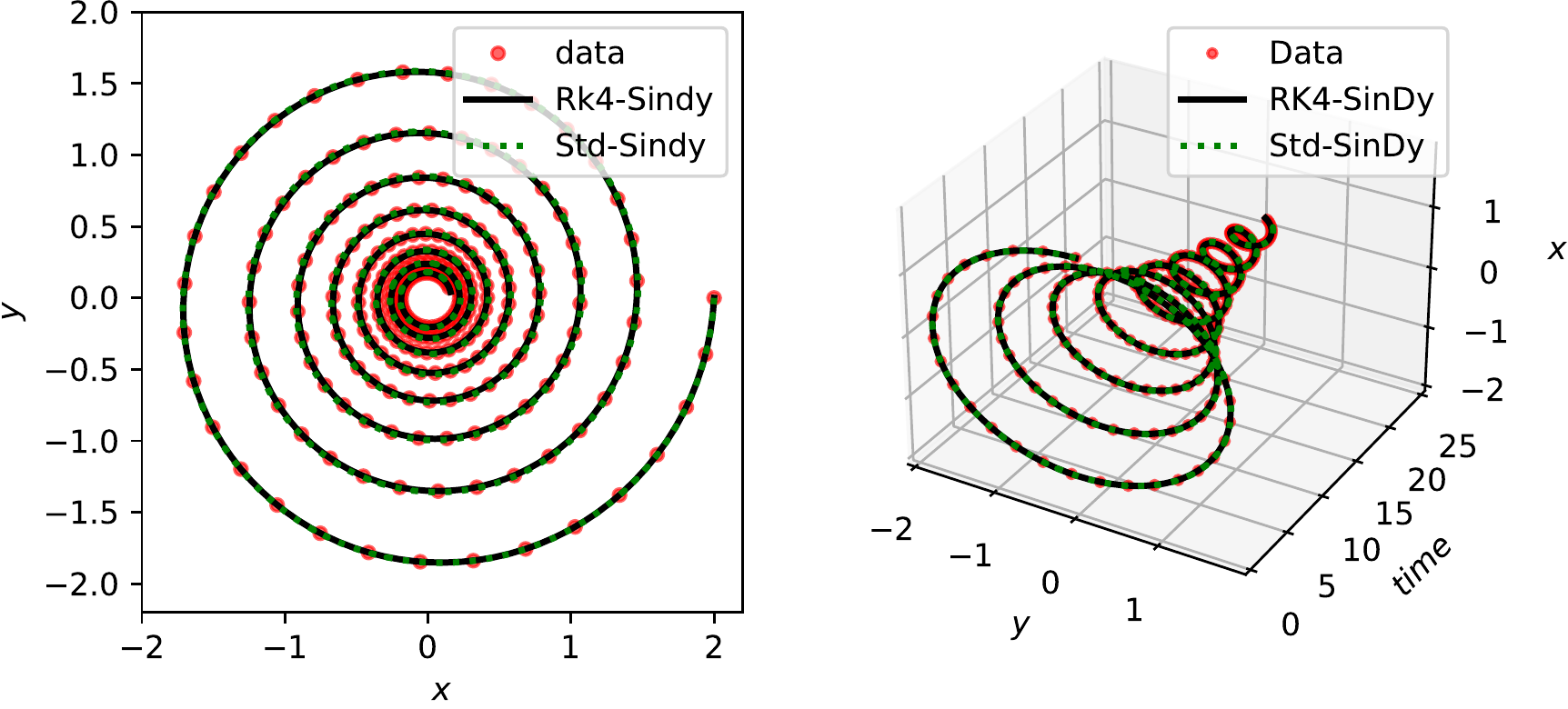}
		\caption{Time step $\dt = 1\cdot 10^{-1}$.}
		\label{fig:linear2d_dt_1}
	\end{subfigure}
	\begin{subfigure}[b]{0.495\textwidth}
		\centering
		\includegraphics[width=1\textwidth]{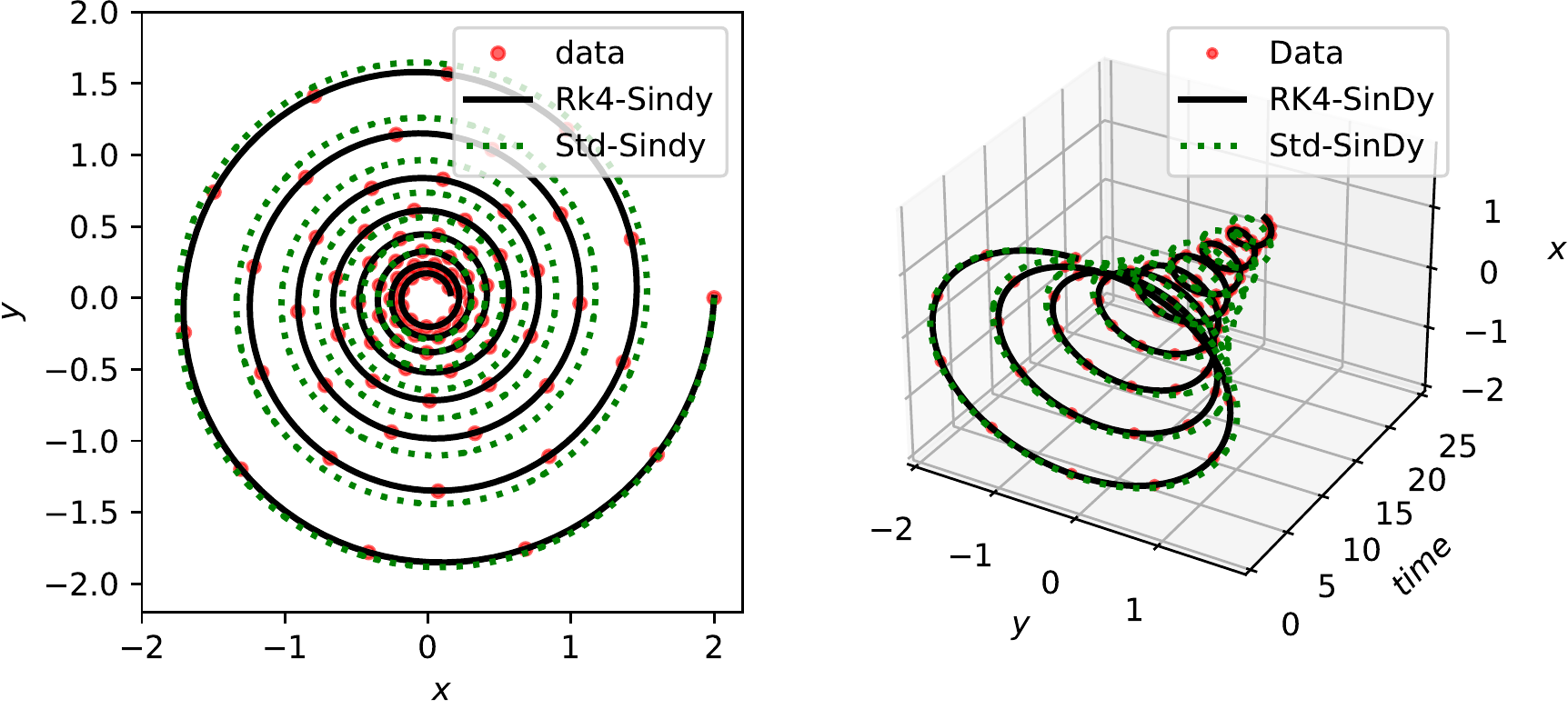}
		\caption{Time step $\dt = 3\cdot 10^{-1}$.}
		\label{fig:linear2d_dt_3}
	\end{subfigure}
	\begin{subfigure}[b]{0.495\textwidth}
		\centering
		\includegraphics[width=1\textwidth]{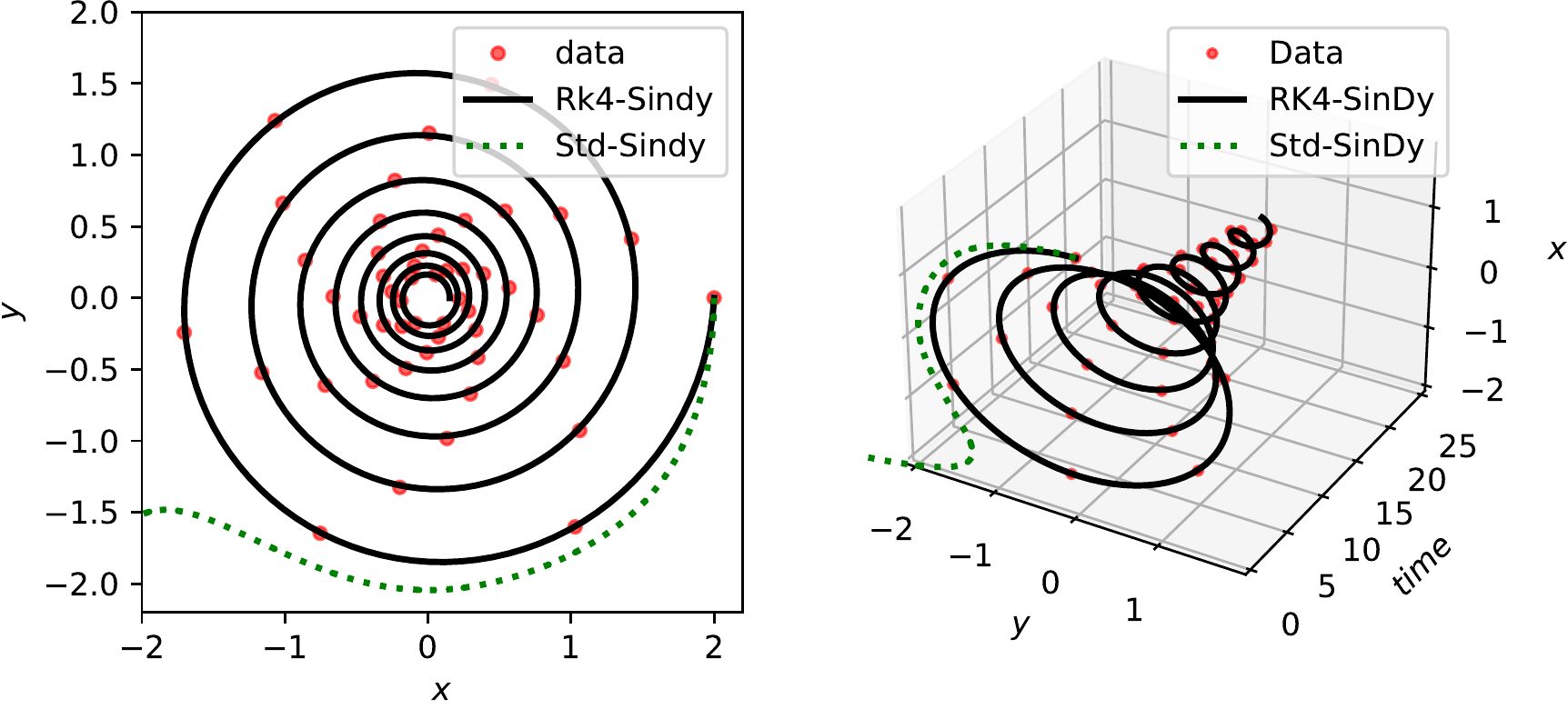}
		\caption{Time step $\dt = 5\cdot 10^{-1}$.}
		\label{fig:linear2d_dt_5}
	\end{subfigure}
	\caption{Linear 2D model: Identified models using data at various regular time-step.}
	\label{fig:linear2D_dt}
\end{figure}

The results are shown in \Cref{fig:linear2D_dt} and \Cref{tab:linear2D_dt}. We notice that \rksindy~is impressively robust with the variation in time-step as compared to \stdsindy, and discovers the governing equations accurately. We also emphasis that  for large time-steps, \stdsindy~fails to capture dynamics; in fact, for a time-step $\dt=5\cdot 10^{-1}$, \stdsindy~even yields unstable models, see \Cref{fig:linear2d_dt_5}.

\renewcommand{\arraystretch}{2.4}
\begin{table}[H]
	\begin{center}
		\rowcolors{1}{}{lightgray}
		\begin{tabular}{|c|c|c|} \hline
			Time step & \bred{\large\rksindy} & \bred{\large\stdsindy} \\ \hline
			$1\cdot10^{-2}$ & $\begin{aligned}
				\dot \bx(t)  & = -0.100 \bx(t) + 2.000\by(t)\\
				\dot \by(t) &= -2.001\bx(t) -0.100\by(t)
			\end{aligned}$ & $\begin{aligned}
				\dot \bx(t)  & = -0.100 \bx(t) + 2.000\by(t)\\
				\dot \by(t) &= -2.000\bx(t) -0.100\by(t)
			\end{aligned}$ \\ \hline
			$1\cdot10^{-1}$ & $\begin{aligned}
				\dot \bx(t)  & = -0.100 \bx(t) + 2.001\by(t)\\
				\dot \by(t) &= -2.001\bx(t) -0.100\by(t)
			\end{aligned}$ & $\begin{aligned}
				\dot \bx(t)  & = -0.098 \bx(t) + 1.987\by(t)\\
				\dot \by(t) &= -1.988\bx(t) -0.098\by(t)
			\end{aligned}$ \\ \hline
			$3\cdot10^{-1}$ & $\begin{aligned}
				\dot \bx(t)  & = -0.101 \bx(t) + 2.002\by(t)\\
				\dot \by(t) &= -2.002\bx(t) -0.101\by(t)
			\end{aligned}$ & $\begin{aligned}
				\dot \bx(t)  & = -0.078 \bx(t) + 1.884\by(t)\\
				\dot \by(t) &= -1.906\bx(t) -0.084\by(t)
			\end{aligned}$ \\ \hline
			$5\cdot10^{-1}$ & $\begin{aligned}
				\dot \bx(t)  & = -0.103 \bx(t) + 2.011\by(t)\\
				\dot \by(t) &= -2.011\bx(t) -0.103\by(t)
			\end{aligned}$ & {\scriptsize $\begin{aligned}
					\dot \bx(t)  & = 1.688\by(t)\\
					\dot \by(t) &= -1.864\bx(t) -0.123\bx(t)^2 \\
					&\hspace{-0.6cm}- 0.146\bx(t)^2\by(t) + 0.115\bx(t)^4\by(t) \\
					&\hspace{-0.6cm}+ 0.133\bx(t)^3\by(t)
				\end{aligned}$} \\ \hline
		\end{tabular}
	\end{center}
	\caption{Linear 2D model: The discovered governing equations using \rksindy~and \stdsindy~are reported for different regular time-steps at which data are collected.}
	\label{tab:linear2D_dt}
\end{table}

\begin{figure}[H]
	\centering
	\begin{subfigure}[b]{0.495\textwidth}
		\centering
		\includegraphics[width=1\textwidth, height = 3cm]{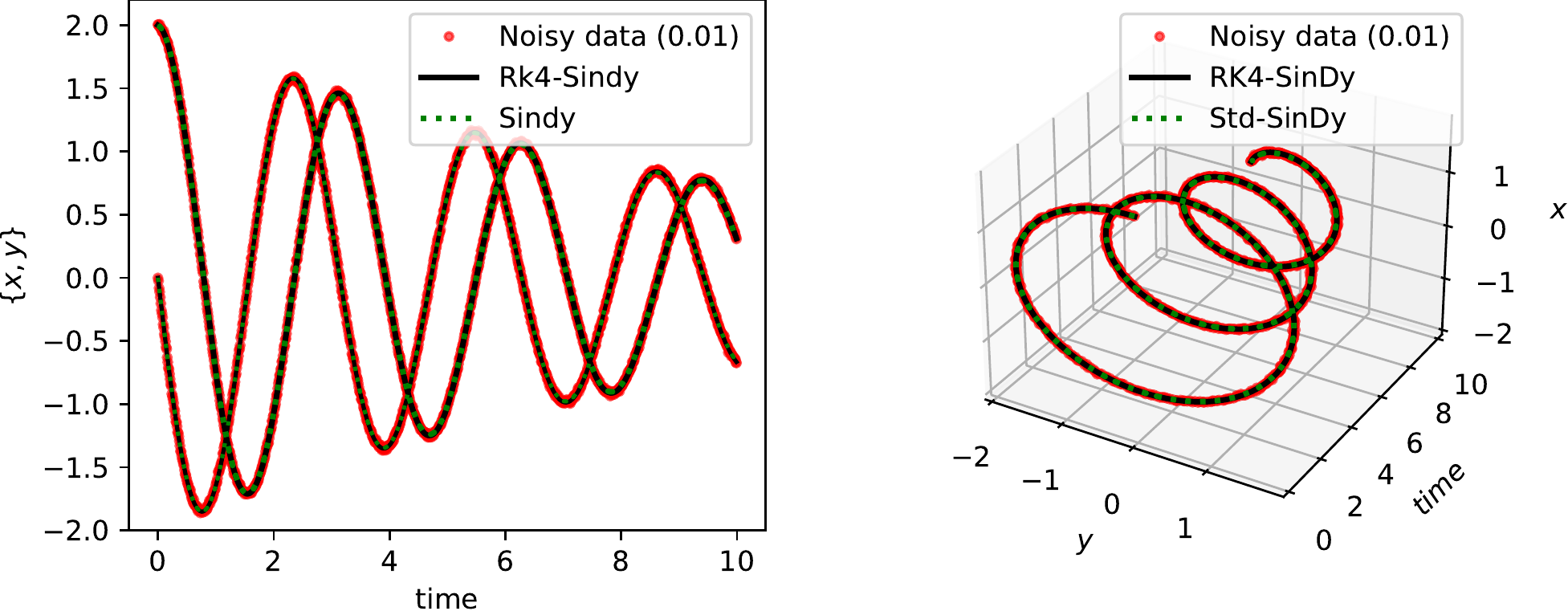}
		\caption{Noise level $\sigma = 1 \cdot 10^{-2}$.}
		\label{fig:linear2d_noise_01}
	\end{subfigure}
	\begin{subfigure}[b]{0.495\textwidth}
		\centering
		\includegraphics[width=1\textwidth, height = 3cm]{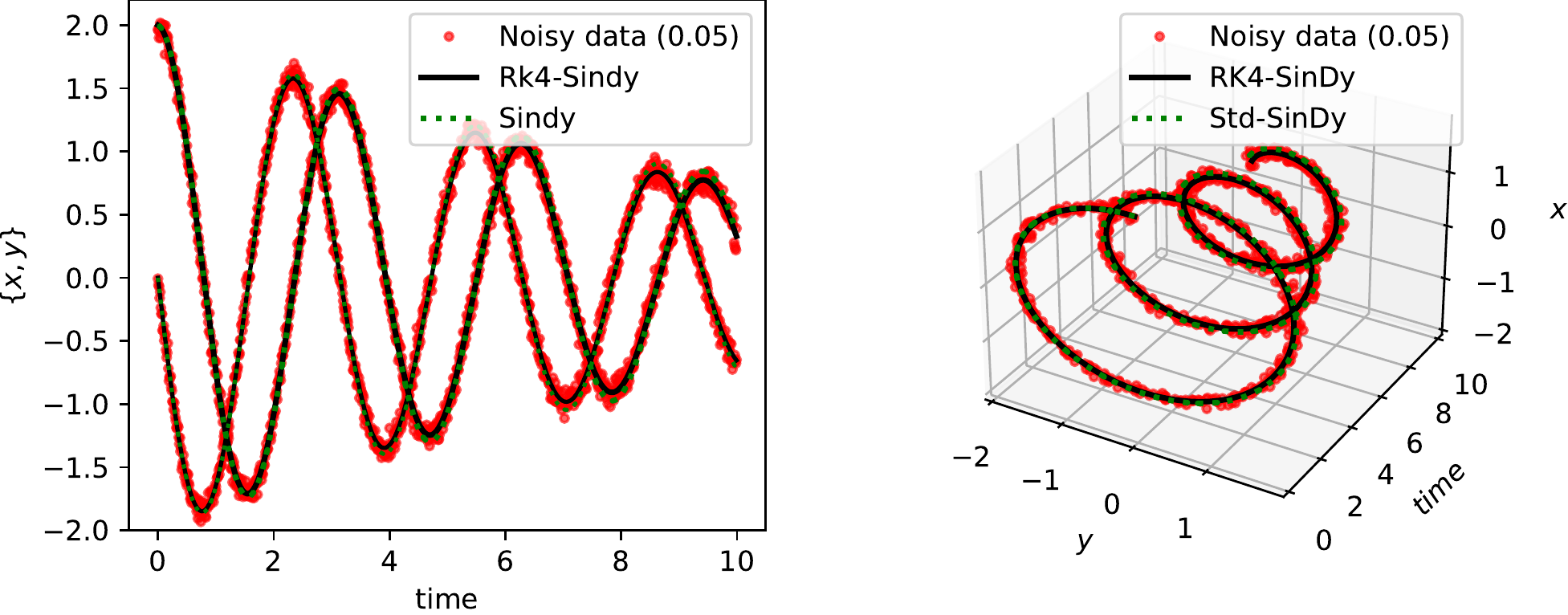}
		\caption{Noise level $\sigma = 5 \cdot 10^{-2}$.}
		\label{fig:linear2d_noise_05}
	\end{subfigure}
	\begin{subfigure}[b]{0.495\textwidth}
		\centering
		\includegraphics[width=1\textwidth, height = 3cm]{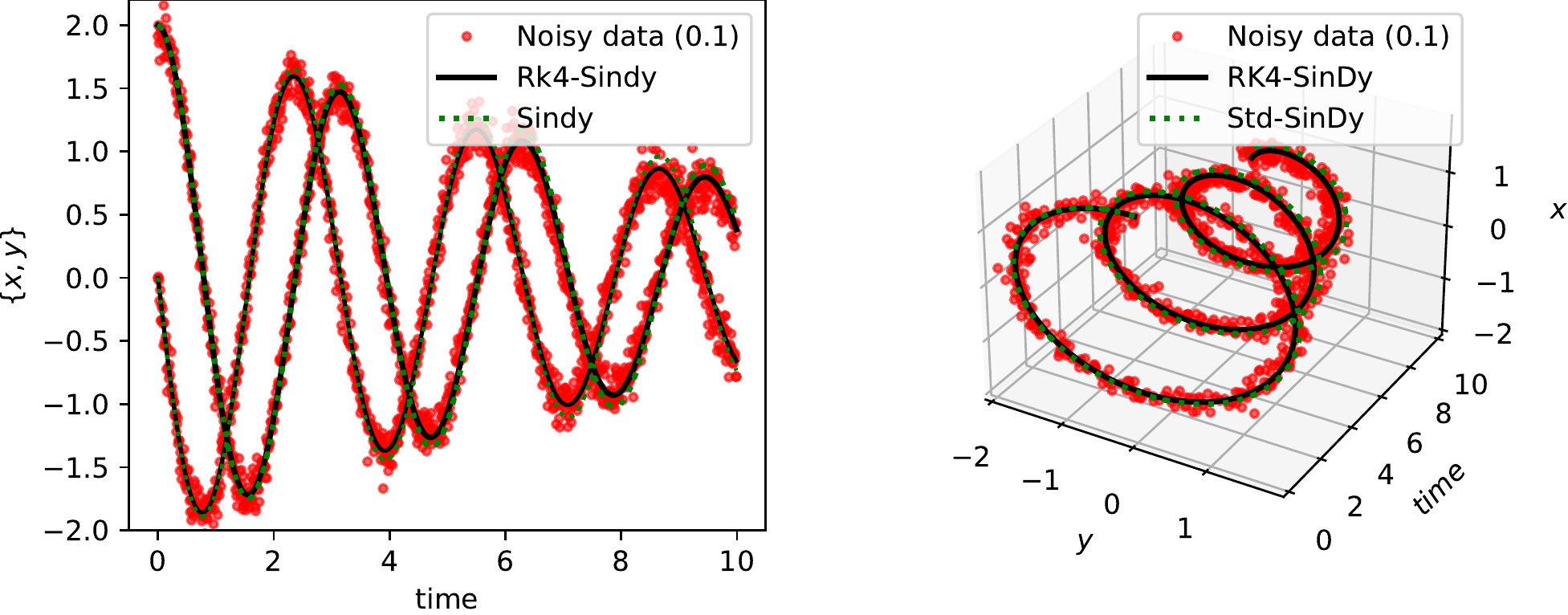}
		\caption{Noise level $\sigma = 1 \cdot 10^{-1}$.}
		\label{fig:linear2d_noise_1}
	\end{subfigure}
	\begin{subfigure}[b]{0.495\textwidth}
		\centering
		\includegraphics[width=1\textwidth, height = 3cm]{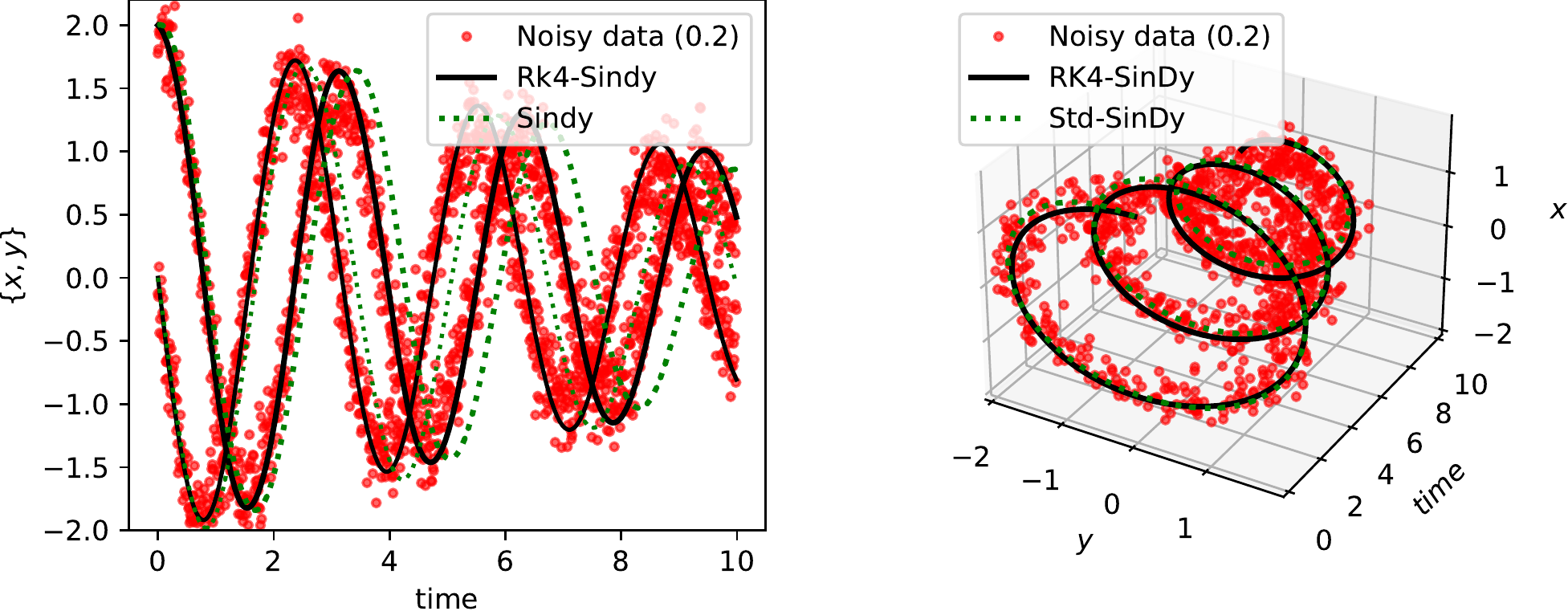}
		\caption{Noise level $\sigma = 2 \cdot 10^{-1}$.}
		\label{fig:linear2d_noise_2}
	\end{subfigure}
	\caption{Linear 2D model: The transient responses of discovered models using corrupted data are compared.}
	\label{fig:linear2d_noise}
\end{figure}

Next, we study the performance of both methodologies under corrupted data. We corrupt the measurement data by adding zero-mean Gaussian white noise of different variances. We present the results in \Cref{fig:linear2d_noise} and \Cref{tab:linear2D_noise} and notice that \rksindy~can discover better quality sparse parsimonious models as compared to \stdsindy~even under significantly corrupted data. It is predominately visible in \Cref{fig:linear2d_noise_2}. 

\renewcommand{\arraystretch}{2.4}
\begin{table}[h]
	\begin{center}
		\rowcolors{1}{}{lightgray}
		\begin{tabular}{|c|c|c|} \hline
			Noise level & \bred{\large\rksindy} & \bred{\large\stdsindy} \\ \hline
			$1\cdot10^{-2}$  & $\begin{aligned}
				\dot \bx(t)  & = -0.099 \bx(t) + 1.999\by(t)\\
				\dot \by(t) &= -2.000\bx(t) - 0.101\by(t)
			\end{aligned}$ & $\begin{aligned}
				\dot \bx(t)  & = -0.102 \bx(t) + 1.999\by(t)\\
				\dot \by(t) &= -2.002\bx(t) -0.101\by(t)
			\end{aligned}$ \\ \hline
			$5\cdot10^{-2}$ & $\begin{aligned}
				\dot \bx(t)  & = -0.095 \bx(t) + 1.999\by(t)\\
				\dot \by(t) &= -1.995\bx(t) -0.105\by(t)
			\end{aligned}$ & $\begin{aligned}
				\dot \bx(t)  & = -0.078 \bx(t) + 2.001\by(t)\\
				\dot \by(t) &= -1.995\bx(t) -0.105\by(t)
			\end{aligned}$ \\ \hline
			$1\cdot10^{-1}$ & $\begin{aligned}
				\dot \bx(t)  & = -0.091 \bx(t) + 1.985\by(t)\\
				\dot \by(t) &= -1.997\bx(t) -0.103\by(t)
			\end{aligned}$ & $\begin{aligned}
				\dot \bx(t)  & = -0.076 \bx(t) + 1.969\by(t)\\
				\dot \by(t) &= -2.008\bx(t) -0.095\by(t)
			\end{aligned}$ \\ \hline
			$2\cdot10^{-1}$ & {\scriptsize $\begin{aligned}
					\dot \bx(t)  & = -0.177 \bx(t) + 2.053\by(t) \\
					&- 0.063\bx^2\by + 0.059\bx\by^2\\
					\dot \by(t) &= -1.960\bx(t)
				\end{aligned}$} & {\scriptsize $\begin{aligned}
					\dot \bx(t)  & =  - 0.173\bx(t) + 1.950\by(t)  -0.056\by(t)^2 \\&+ 0.059\bx(t)^3  - 0.079\bx(t)^2\by  + 0.095\\
					\dot \by(t) &= -2.005\bx(t) -0.095\by(t)  + 0.069\bx(t)\by(t) \\ &+ 0.062\bx(t)^3 + 0.060\bx(t)\by(t)^2
				\end{aligned}$ }\\ \hline
		\end{tabular}
	\end{center}
	\caption{Linear 2D model: The discovered governing equations, by employing \rksindy~and \stdsindy, are reported. In this scenario, the measurement data  are corrupted using zero-mean Gaussian white noise of different variances.}
	\label{tab:linear2D_noise}
\end{table}


\subsubsection{Cubic damped oscillator}
\begin{figure}[H]
	\centering
	\begin{subfigure}[b]{0.495\textwidth}
		\centering
		\includegraphics[width=1\textwidth]{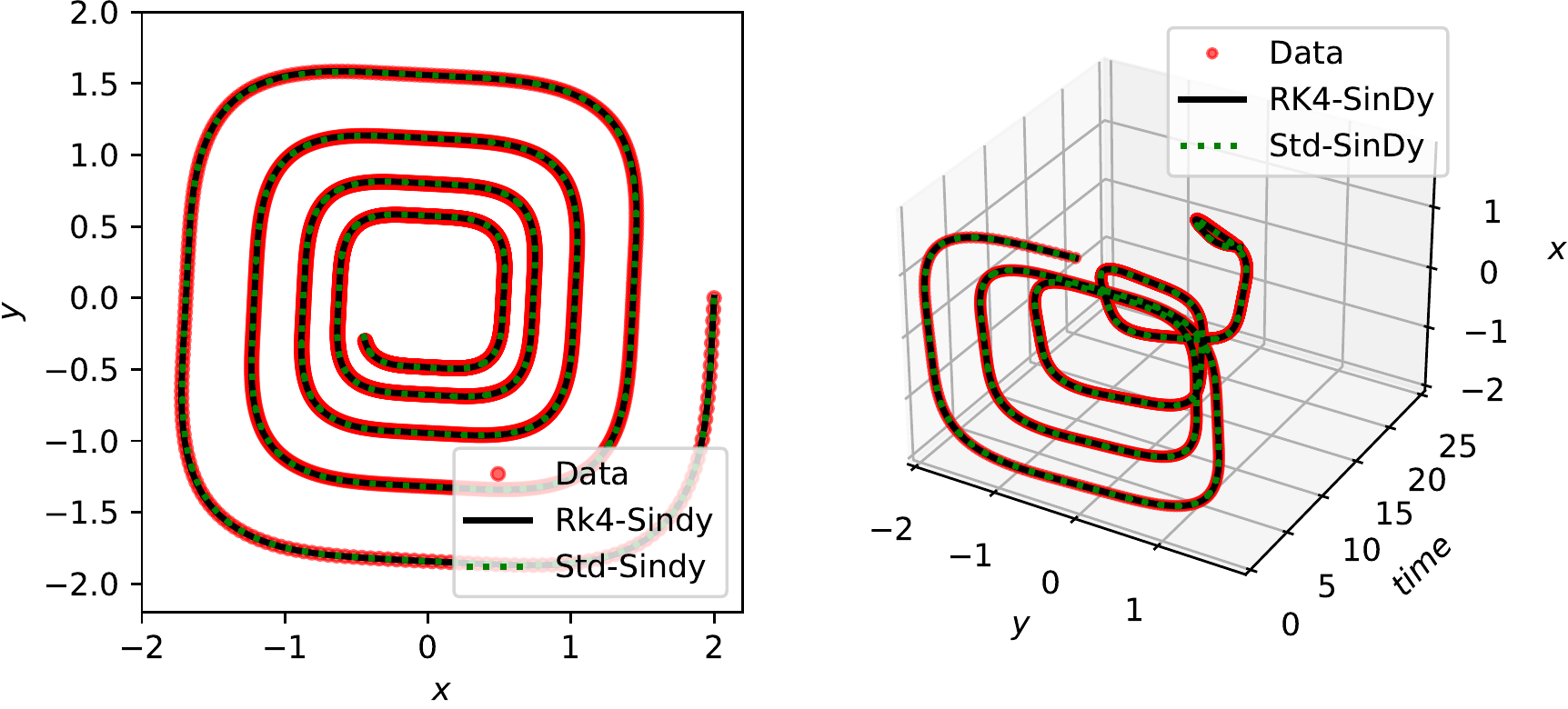}
		\caption{Time step $\dt = 5 \cdot 10^{-3}$.}
		\label{fig:cubic2d_dt_005}
	\end{subfigure}
	\begin{subfigure}[b]{0.495\textwidth}
		\centering
		\includegraphics[width=1\textwidth]{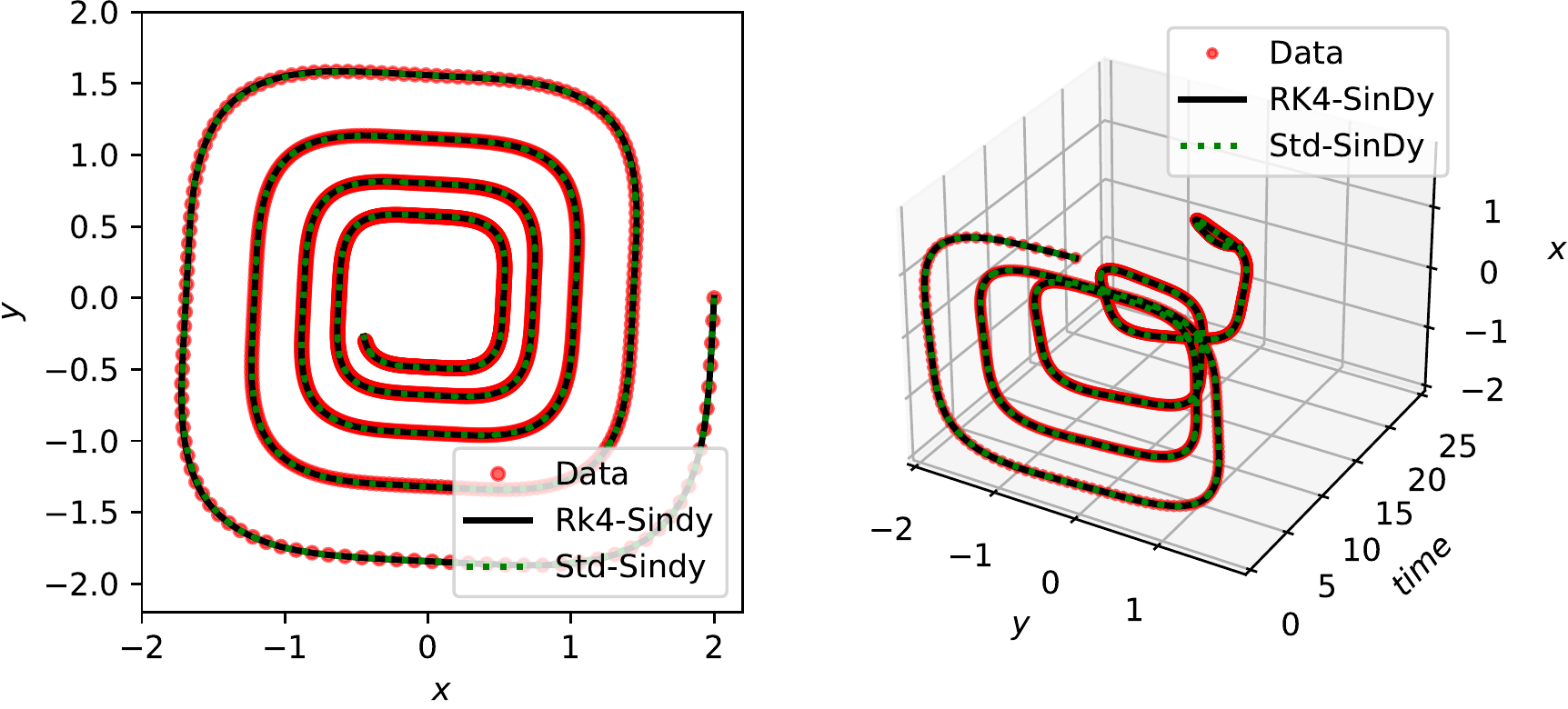}
		\caption{Time step $\dt = 1\cdot 10^{-2}$.}
		\label{fig:cubic2d_dt_01}
	\end{subfigure}
	\begin{subfigure}[b]{0.495\textwidth}
		\centering
		\includegraphics[width=1\textwidth]{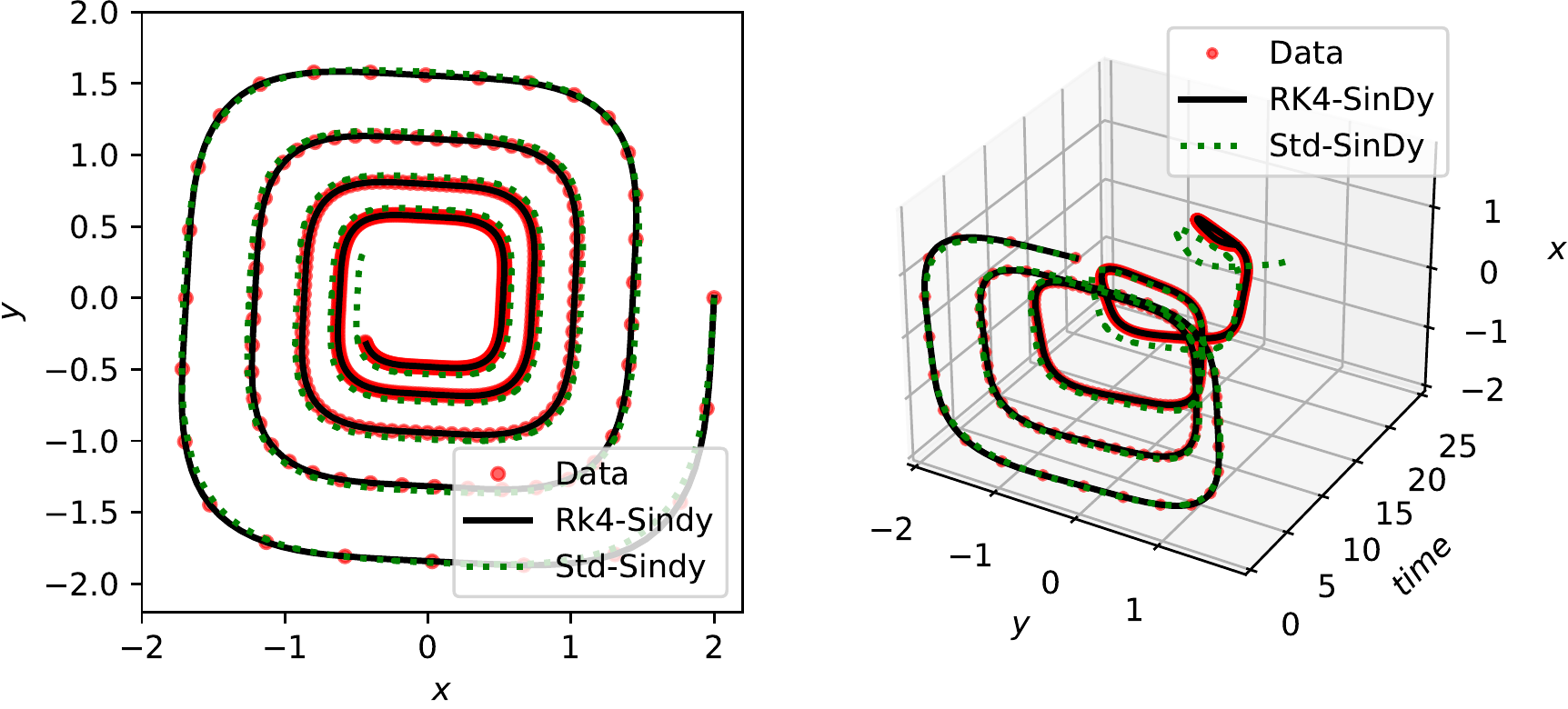}
		\caption{Time step $\dt = 5\cdot 10^{-2}$.}
		\label{fig:cubic2d_dt_05}
	\end{subfigure}
	\begin{subfigure}[b]{0.495\textwidth}
		\centering
		\includegraphics[width=1\textwidth]{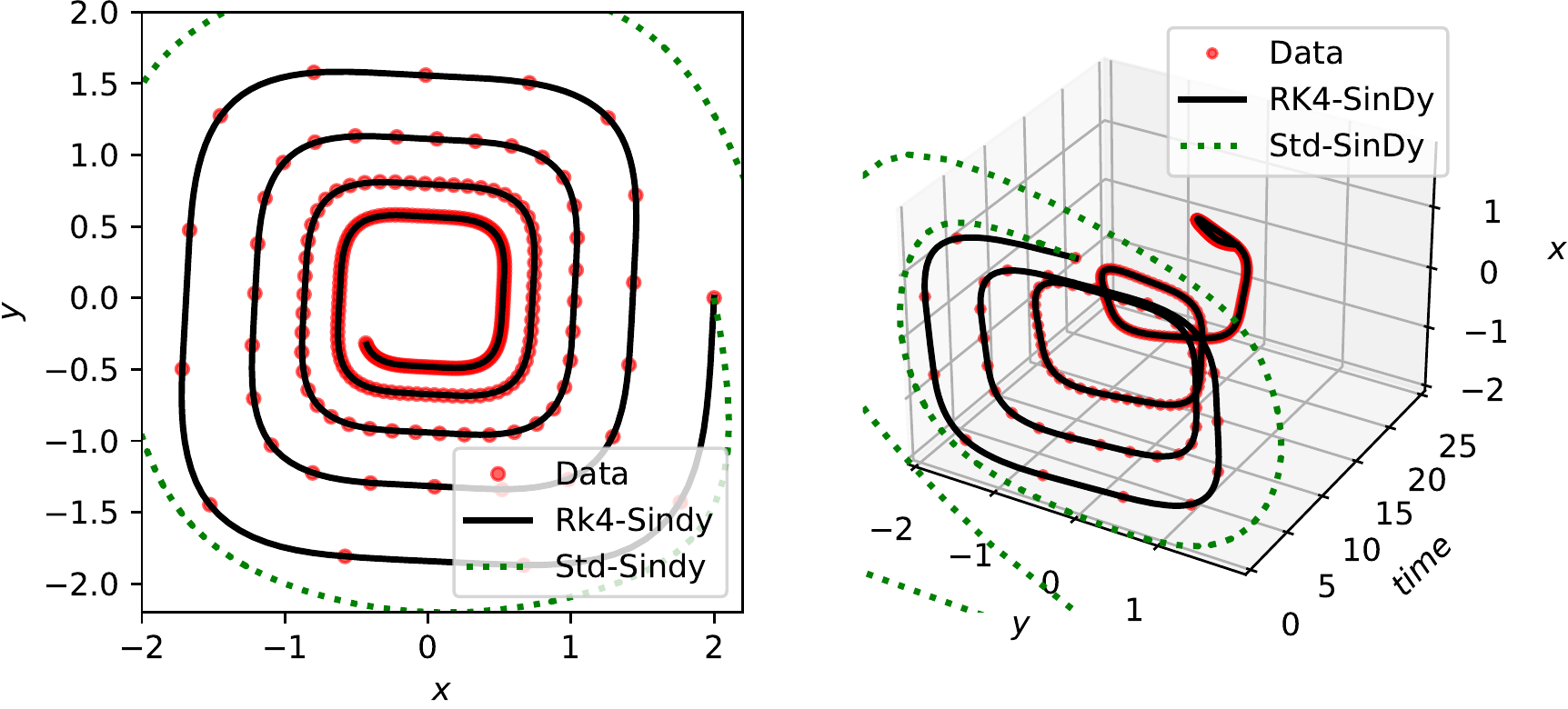}
		\caption{Time step $\dt = 1\cdot 10^{-1}$.}
		\label{fig:cubic2d_dt_1}
	\end{subfigure}
	\caption{Cubic 2D model: A comparison of the transient responses of discovered models using data at different regular time-steps.}
	\label{fig:cubic2D_dt}
\end{figure}

Next, we consider a cubic damped oscillator, governed by
\begin{equation}
	\begin{aligned}
		\dot\bx(t) &= -0.1\bx(t)^3 + 2.0\by(t)^3,\\
		\dot\bx(t) &= -2.0\bx(t)^3 - 0.1\by(t)^3.
	\end{aligned}
\end{equation}
Like the linear case, we aim at discovering the governing equation using measurement data. We repeat the study done in the previous example using different regular time-steps. We report the quality of discovered models using \rksindy~and \stdsindy~in \Cref{fig:cubic2D_dt} and \Cref{tab:cubic2D_dt}. We observe that \rksindy~successfully discovers the governing equations quite accurately, whereas \stdsindy~struggles to identify the governing equations when measurements data are collected at a larger time-step. It simply fails to obtain a stable model for a time-step $\dt = 0.1$. It showcases the robustness of \rksindy~to discover interpretable models even when data are collected sparsely.

\begin{table}[h]
	\begin{center}
		\rowcolors{1}{}{lightgray}
		\begin{tabular}{|c|c|c|} \hline
			Time step & \bred{\large\rksindy} & \bred{\large\stdsindy} \\ \hline
			$5\cdot10^{-3}$  & $\begin{aligned}
				\dot \bx(t)  & = -0.099 \bx(t)^3 + 1.996\by(t)^3\\
				\dot \by(t) &= -1.997\bx(t)^3 -0.100\by(t)^3
			\end{aligned}$ & $\begin{aligned}
				\dot \bx(t)  & = -0.099 \bx(t)^3 + 1.995\by(t)^3\\
				\dot \by(t) &= -1.996\bx(t)^3 -0.099\by(t)^3
			\end{aligned}$ \\ \hline
			$1\cdot10^{-2}$ & $\begin{aligned}
				\dot \bx(t)  & = -0.099 \bx(t)^3 + 1.995\by(t)^3\\
				\dot \by(t) &= -1.997\bx(t)^3 -0.100\by(t)^3
			\end{aligned}$ & $\begin{aligned}
				\dot \bx(t)  & = -0.100 \bx(t)^3 + 1.994\by(t)^3\\
				\dot \by(t) &= -1.996\bx(t)^3 -0.099\by(t)^3
			\end{aligned}$ \\ \hline
			$5\cdot10^{-2}$ & $\begin{aligned}
				\dot \bx(t)  & = -0.100 \bx(t)^3 + 1.995\by(t)^3\\
				\dot \by(t) &= -1.997\bx(t)^3 -0.100\by(t)^3
			\end{aligned}$ & {\scriptsize$\begin{aligned}
					\dot \bx(t)  & = -0.092 \bx(t)^3 + 2.002\by(t)^3 \\
					&+ 0.076\bx^4\by - 0.107\bx^2\by^3\\
					\dot \by(t) &= -1.981\bx(t)^3 -0.092\by(t)^3 \\
					&+ 0.078\bx^3\by^2 - 0.068\bx\by^4
				\end{aligned}$ }\\ \hline
			$1\cdot10^{-1}$ & $\begin{aligned}
				\dot \bx(t)  & = -0.103 \bx(t) + 2.000\by(t)\\
				\dot \by(t) &= -2.001\bx(t) -0.098\by(t)
			\end{aligned}$ & {\scriptsize$\begin{aligned}
					\dot \bx(t)  & = 0.090\bx(t) - 0.097\bx(t)^2 -0.463\bx(t)^3 \\ & + \cdots + 0.381\bx(t)^3\by(t)^2 -0.258\bx(t)\by(t)^4\\
					\dot \by(t)& = 0.100\bx(t) + 0.104\bx(t)^2 + 0.051\bx(t)\by(t) \\ & + \cdots + 0.381\bx(t)^3\by(t)^2 -0.258\bx(t)\by(t)^4\\
				\end{aligned}$ } \\ \hline
		\end{tabular}
	\end{center}
	\caption{Cubic 2D model: The table reports the discovered governing equations by employing \rksindy~and \stdsindy.}
	\label{tab:cubic2D_dt}
\end{table}

\subsection{Fitz-Hugh Nagumo model}
\begin{figure}[!htb]
	\centering
	\begin{subfigure}[b]{0.495\textwidth}
		\centering
		\includegraphics[width=1\textwidth]{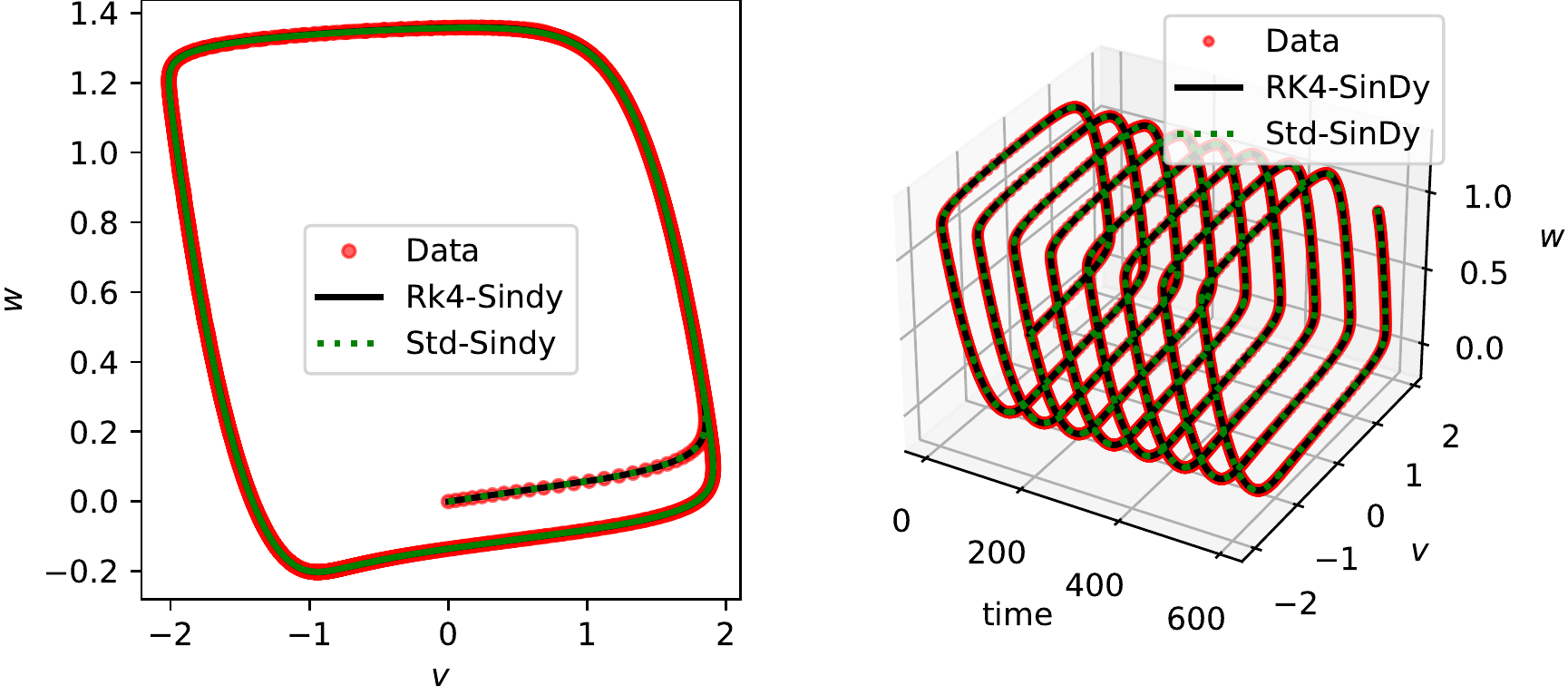}
		\caption{Time step $\dt = 1.0 \cdot 10^{-1}$.}
		\label{fig:FHN_dt_01}
	\end{subfigure}
	\begin{subfigure}[b]{0.495\textwidth}
		\centering
		\includegraphics[width=1\textwidth]{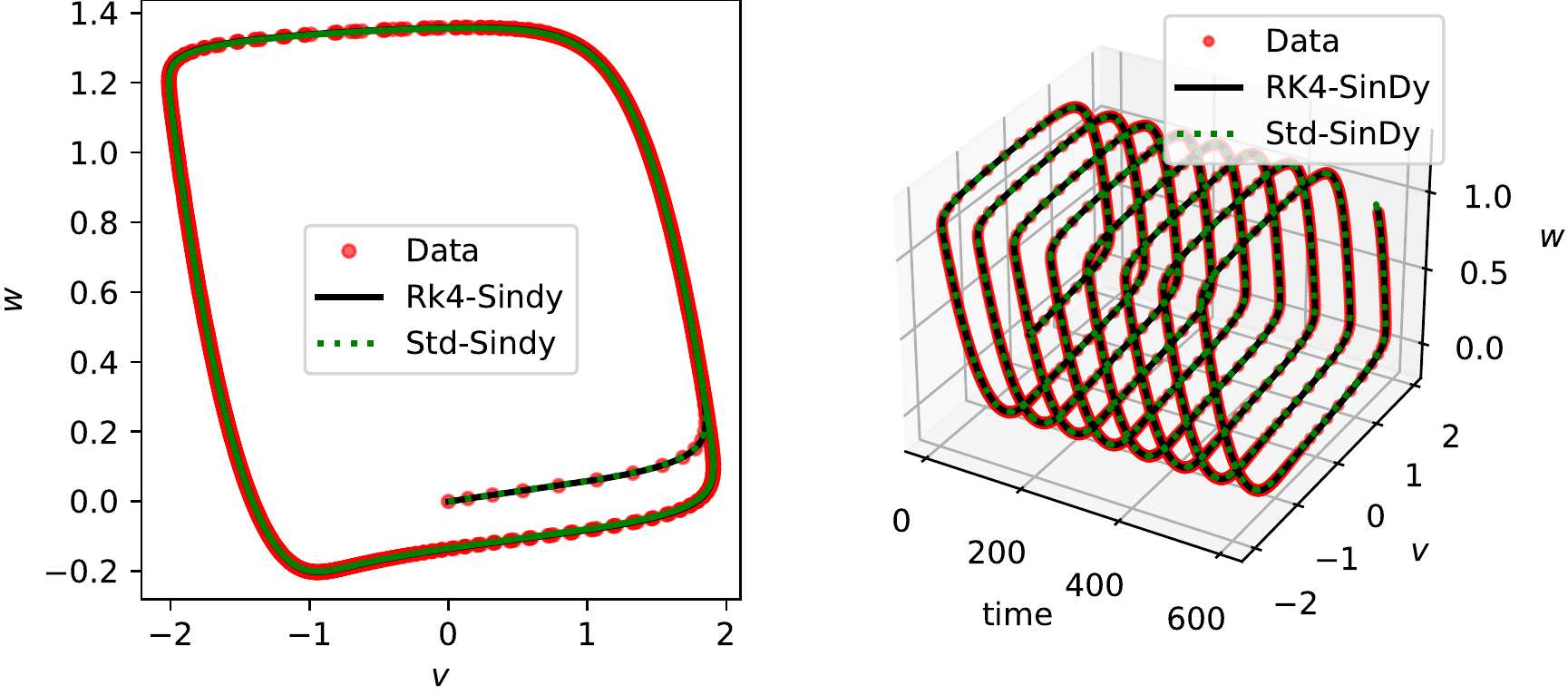}
		\caption{Time step $\dt = 2.5\cdot 10^{-1}$.}
		\label{fig:FHN_dt_025}
	\end{subfigure}
	\begin{subfigure}[b]{0.495\textwidth}
		\centering
		\includegraphics[width=1\textwidth]{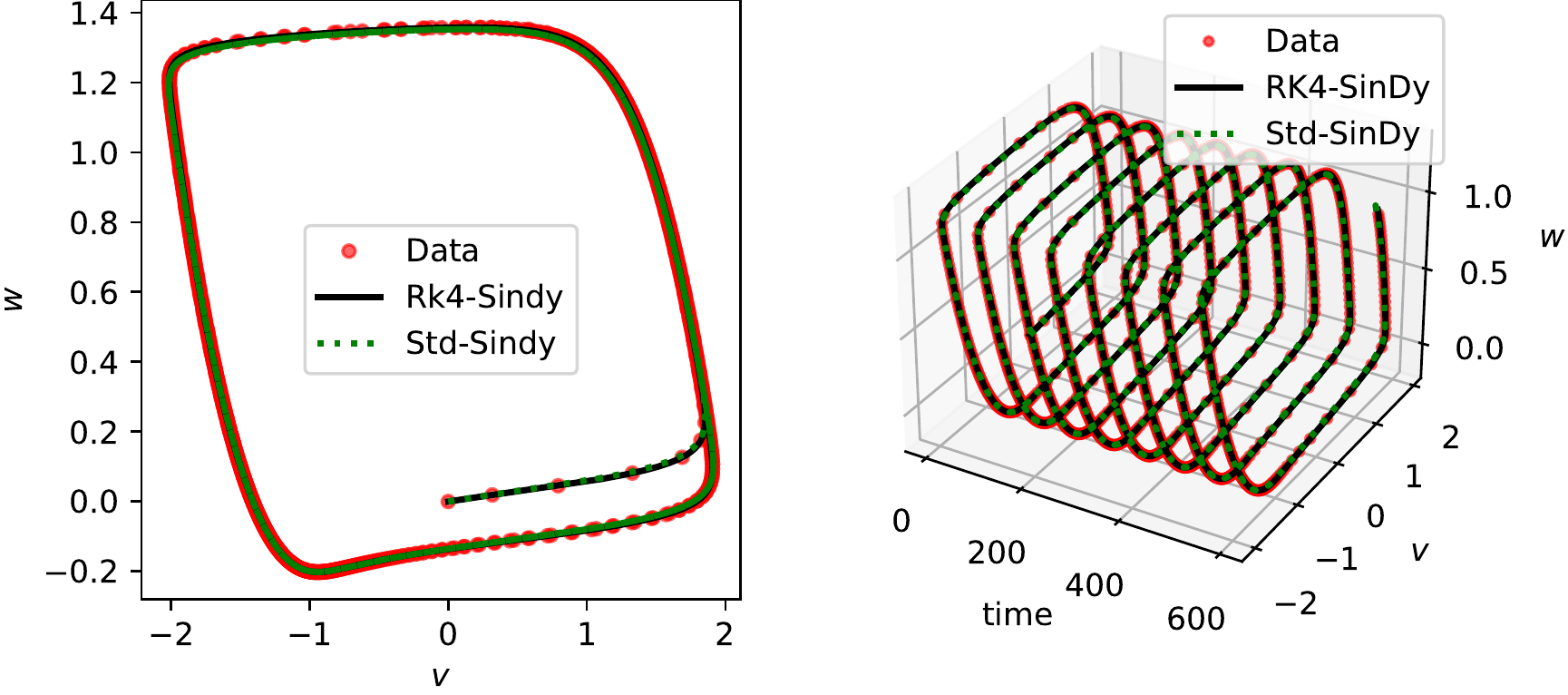}
		\caption{Time step $\dt = 5.0\cdot 10^{-1}$.}
		\label{fig:FHN_dt_05}
	\end{subfigure}
	\begin{subfigure}[b]{0.495\textwidth}
		\centering
		\includegraphics[width=1\textwidth]{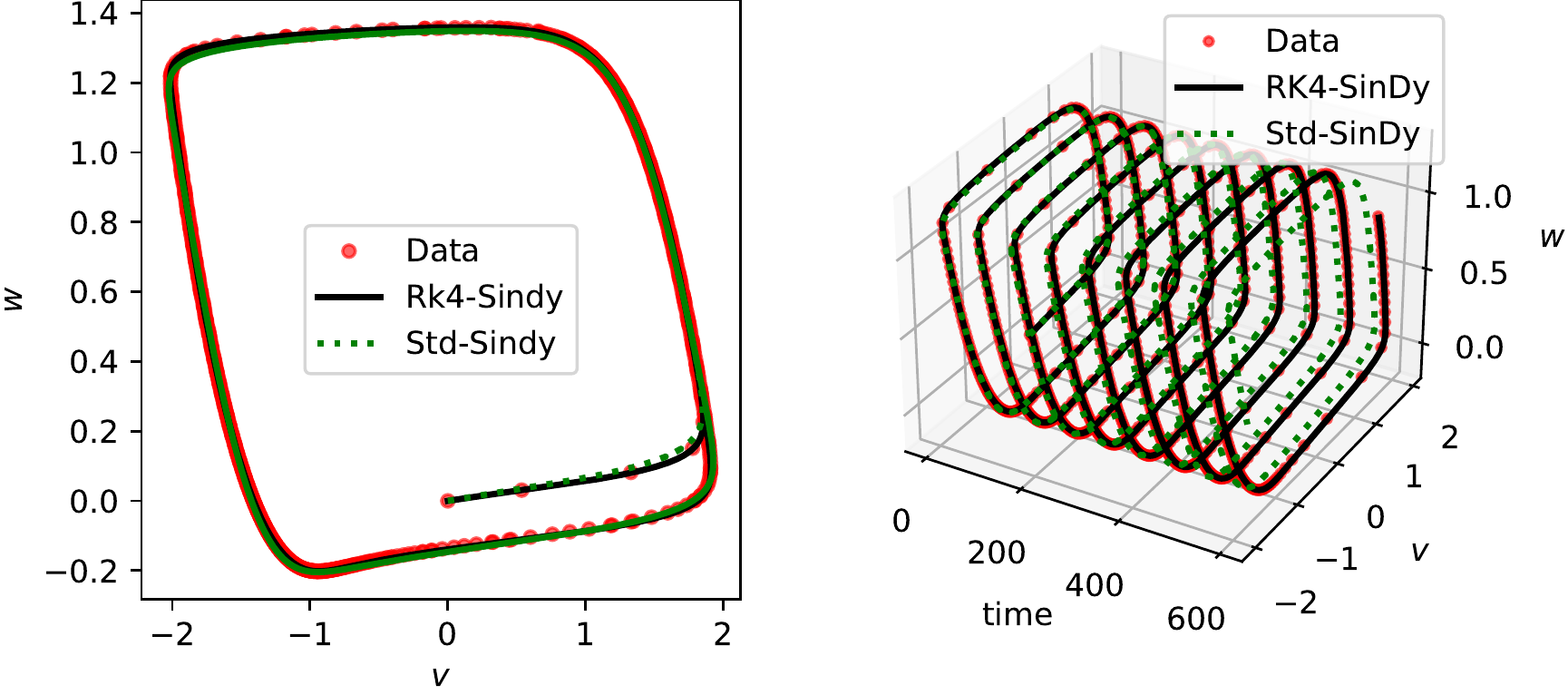}
		\caption{Time step $\dt = 7.5\cdot 10^{-1}$.}
		\label{fig:FHN_dt_075}
	\end{subfigure}
	\caption{FHN model: A comparison of the transient responses of the discovered differential equations using data collected at different regular time-steps.}
	\label{fig:FHN_dt}
\end{figure}

Here, we explore discovery of the nonlinear Fitz-Hugh Nagumo (FHN) model that describes the activation and deactivation of neurons in a simplistic way \cite{fitzhugh1955mathematical}. The governing equations are:
\begin{equation}
	\begin{aligned}
		\bv(t) &= \bv(t) - \bw(t) - \dfrac{1}{3}\bv(t)^3 + 0.5,\\
		\bw(t) &= 0.040\bv(t) - 0.028\bw(t) + 0.032.
	\end{aligned}
\end{equation}
We collect the time-history data of $\bv(t)$ and $\bw(t)$ using the zero initial condition. We construct a dictionary containing polynomial terms up to the third degree. We employ \rksindy~(\Cref{algo:procedure1} with $\lambda = 10^{-2}$) and \stdsindy. We discover governing equations by using the data collected between the time interval $\left[0,600\right]$s. We identify models under different conditions, namely, different time-steps at which data are collected. We report the results in \Cref{fig:FHN_dt} and \Cref{tab:FHN_dt}. It can be observed that \rksindy~faithfully discovers the underlying governing equations by picking the correct features from the dictionary and estimates the corresponding coefficients up to $1\%$ accurately. On the other hand, \stdsindy~breaks down when data are taken at a large time-step. 

\begin{table}[!htb]
	\begin{center}
		\rowcolors{1}{}{lightgray}
		\begin{tabular}{|c|c|c|} \hline
			$\dt$ & \bred{\large\rksindy} & \bred{\large\stdsindy} \\ \hline
			$1.0\cdot10^{-1}$  & $\begin{aligned}
				\dot \bv(t)  & =  0.499 + 0.998\bv - 0.998 \bw - 0.333\bv^3\\
				\dot \bw(t) &= 0.032 + 0.040\bv -0.028\bw
			\end{aligned}$ & $\begin{aligned}
				\dot \bv(t)  & =  0.498 + 0.996\bv - 0.996 \bw - 0.332\bv^3\\
				\dot \bw(t) &= 0.032 + 0.040\bv -0.028\bw
			\end{aligned}$ \\ \hline
			$2.5\cdot10^{-1}$ & $\begin{aligned}
				\dot \bv(t)  & =  0.499 + 0.998\bv - 0.998 \bw - 0.333\bv^3\\
				\dot \bw(t) &= 0.032 + 0.040\bv -0.028\bw
			\end{aligned}$ & $\begin{aligned}
				\dot \bv(t)  & =  0.494 + 0.985\bv - 0.989 \bw - 0.328\bv^3\\
				\dot \bw(t) &= 0.032 + 0.040\bv -0.028\bw
			\end{aligned}$ \\ \hline
			$5.0\cdot10^{-1}$ & $\begin{aligned}
				\dot \bv(t)  & =  0.501 + 1.001\bv - 1.001 \bw - 0.334\bv^3\\
				\dot \bw(t) &= 0.032 + 0.040\bv -0.028\bw
			\end{aligned}$ & $\begin{aligned}
				\dot \bv(t)  & =  0.482 + 0.943\bv - 0.959 \bw \\ &- 0.034\bv\bw - 0.311\bv^3 + 0.024\bv\bw^2\\
				\dot \bw(t) &= 0.032 + 0.040\bv -0.028\bw
			\end{aligned}$ \\ \hline
			$7.5\cdot10^{-1}$ & $\begin{aligned}
				\dot \bv(t)  & =  0.502 + 1.001\bv - 1.003 \bw - 0.334\bv^3\\
				\dot \bw(t) &= 0.032 + 0.040\bv -0.027\bw
			\end{aligned}$ & $\begin{aligned}
				\dot \bv(t)  & =  0.459 + 0.816\bv - 0.982\bw \\ &- 0.013\bv^2 + \cdots + 0.131\bv\bw^2 - 0.021\bw^3\\
				\dot \bw(t) &= 0.032 + 0.040\bv -0.028\bw
			\end{aligned}$ \\ \hline
		\end{tabular}
	\end{center}
	\caption{FHN model: Discovered models using data at various time-step using \rksindy~and \stdsindy.}
	\label{tab:FHN_dt}
\end{table}

\subsection{Chaotic Lorenz system}\label{subsec:Lorenz}
\begin{figure}[H]
	\centering
	\includegraphics[width = \textwidth]{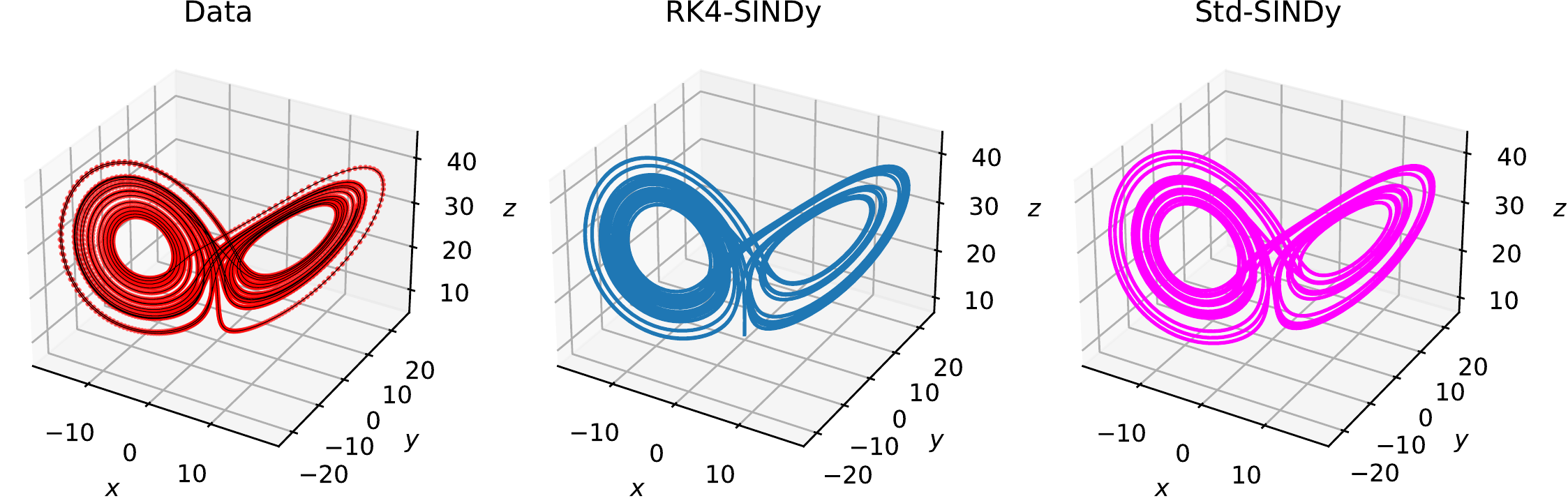}
	\caption{Chaotic Lorenz model: The left figures shows the collected data (in red) and a finely spaced trajectory of the ground truth is shown in black. The middle and right figures shows the trajectories obtained using the discovered models using \rksindy~and \stdsindy, respectively.}
	\label{fig:lorez_noisefree}
\end{figure}

As the next example, we consider the problem of discovering the nonlinear Lorenz model \cite{lorenz1963deterministic}. The dynamics of the chaotic system involves on an attractor and is governed by
\begin{equation}\label{eq:lorenz}
	\begin{aligned}
		\dot\bx(t) &= -10 \bx(t) + 10 \by(t),\\
		\dot \by(t) &= \bx(28-\bz(t)) - \by(t), \\
		\dot\bz(t) &= \bx(t)\by(t) - \tfrac{8}{3}\bz(t).
	\end{aligned}
\end{equation} 
We collect the data by simulating the model from time $t =0$ to $t = 20$ with a time-step of $\dt = 10^{-2}$.  To discover the governing equations using the measurement data, we employ \rksindy~and \stdsindy~with the fixed cutoff parameter $\lambda = 0.5$. However, before employing the methodologies, we perform a normalization step. A reason behind is that the mean value of the variable $\bz$ is large, and the standard deviations of all the three variables is much larger than $1$. Consequently, a dictionary containing polynomial terms would be highly ill-conditioned. To circumvent this, we perform a normalization of data. Ideally, one performs normalization such that the mean and variance of the transformed data are $0$ and $1$. But for this particular example, we normalize such that the interactions between the transformed variables are similar to \eqref{eq:lorenz}. Hence, we propose a transformation as 
\begin{equation}
	\tilde{\bx}(t) := \tfrac{\bx(t)}{8},\quad 	\tilde{\by}(t) := \tfrac{\by(t)}{8}, \quad 	\tilde{\bz}(t) := \tfrac{\bz(t) - 25}{8}.
\end{equation}
Consequently, we obtain a model:
\begin{equation}\label{eq:lorenz_transformed}
	\begin{aligned}
		\dot{\tilde\bx}(t) &= -10 \tilde\bx(t) + 10 \tilde\by(t),\\
		\dot {\tilde\by}(t) &= \tilde\bx(28-8\tilde\bz(t)) - \tilde\by(t), \\
		\dot{\tilde\bz}(t) &= 8\tilde\bx(t)\tilde\by(t) - \tfrac{8}{3}\tilde\bz(t) - \tfrac{25}{3}.
	\end{aligned}
\end{equation}
Notwithstanding, the models \eqref{eq:lorenz} and \eqref{eq:lorenz_transformed} look alike, and the basis features in which dynamics of both models lie are the same except a constant. However, the beauty of the model \eqref{eq:lorenz_transformed} or the transformed data is that the data becomes well-conditioned, hence the dictionary containing polynomial features. Next, we discover models by employing \rksindy~and \stdsindy~using the transformed data. For this, we construct a dictionary with polynomial nonlinearities up to degrees $3$.  We observe the result in  \Cref{fig:lorez_noisefree} and \Cref{tab:lorez_noisefree}. We note that both methods identify correct features from the dictionary with coefficients that are close to the ground truth, but \rksindy~model coefficients are relatively closer to the ground-truth ones. It is also worthwhile to note that the coefficients of the obtained \rksindy~model are only $0.01\%$ off to the ground-truth, but the dynamics still seem quite different, see \Cref{fig:lorez_noisefree}. A reason behind this is the highly chaotic behavior of the dynamics. As a result, a tiny deviation in the coefficients can significantly impact the transient behavior in an absolute sense; however, the dynamics on an attractor are perfectly captured. 

\begin{figure}[!htb]
	\centering
	\includegraphics[width = 0.7\textwidth]{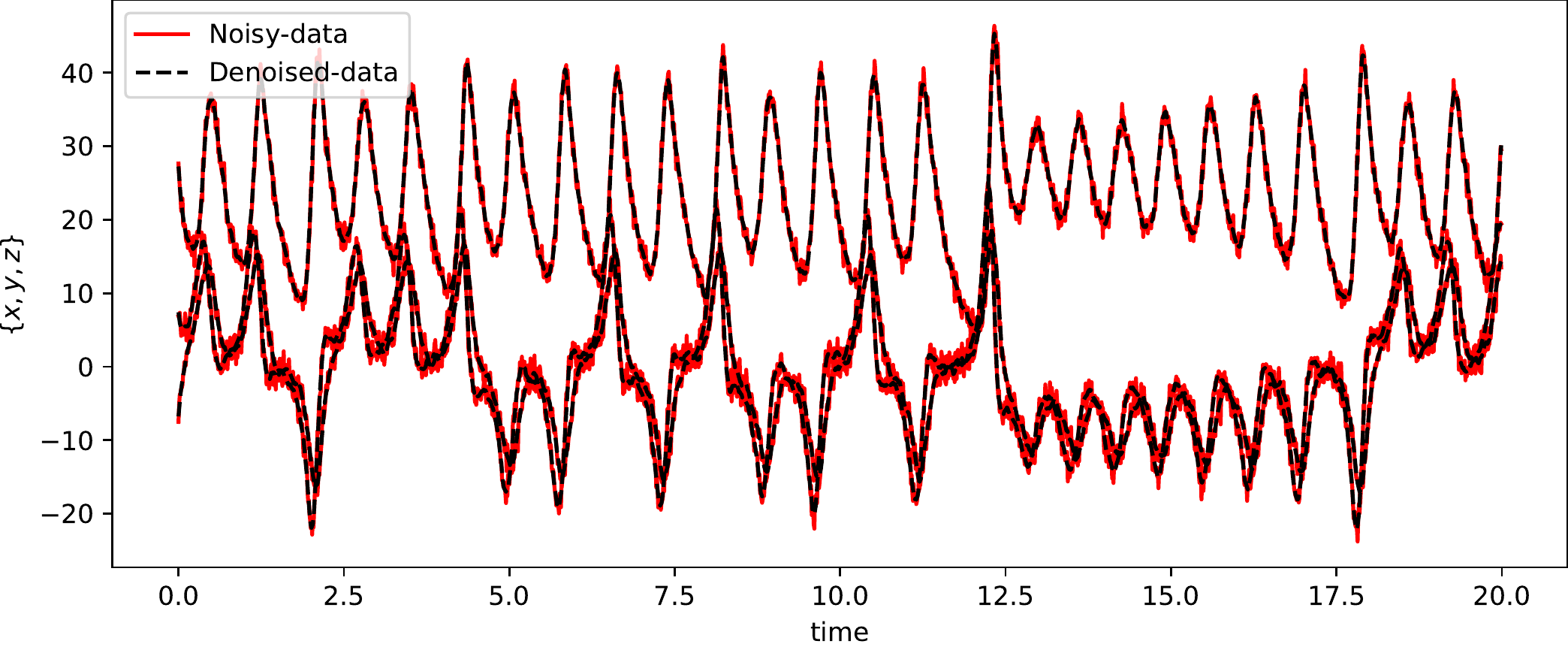}
	\caption{Chaotic Lorenz model: The figure shows the noisy measurements of $\{\bx,\by,\bz\}$ that are corrupted by adding zero-mean Gaussian noise of variance one. It also shows the denoised signal done using a Savitzky-Golay filter \cite{savitzky1964smoothing}.}
	\label{fig:lorez_denoise}
\end{figure}

\begin{figure}[!htb]
	\centering
	\includegraphics[width = \textwidth]{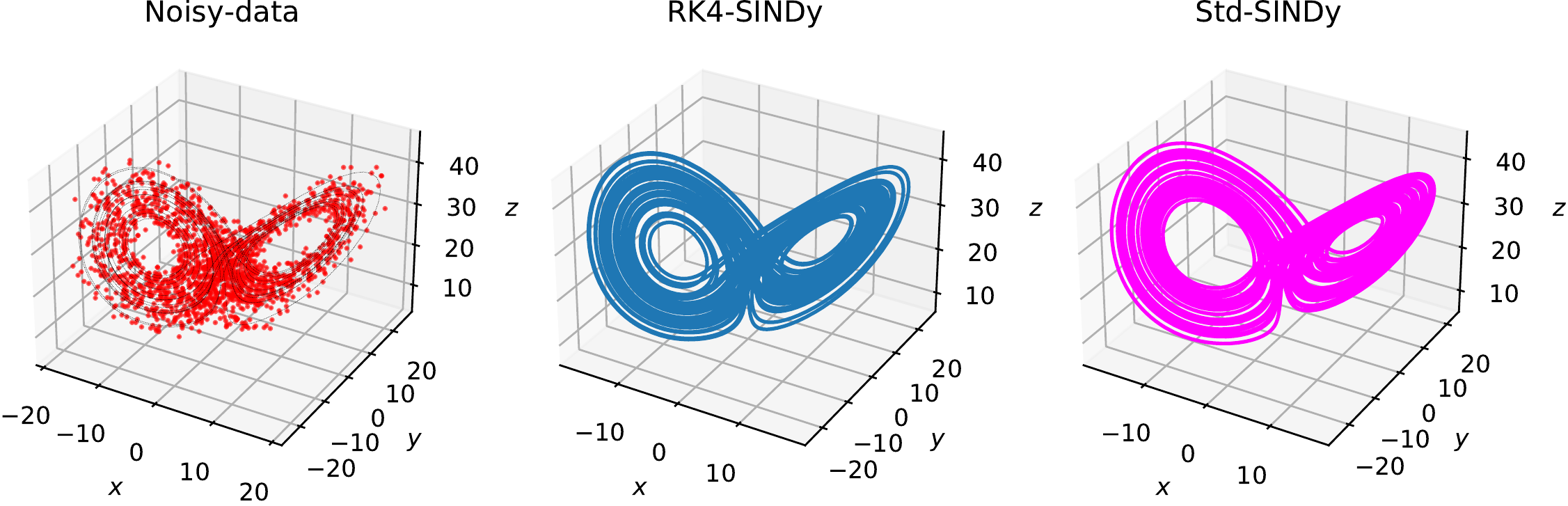}
	\caption{Chaotic Lorenz model: The left figure shows the collected noisy data (in red), and a continuous trajectory of the ground truth is shown in light black. We have added a Gaussian white noise of mean zero and variance one. The middle and right figures present the transient trajectories obtained using the discovered models using \rksindy~and \stdsindy, respectively, and show that the dynamics of the discovered models are intact on an attractor.}
	\label{fig:lorez_noise}
\end{figure}

\begin{table}[H]
	\caption{Chaotic Lorenz model: Discovered governing equations using \rksindy~and \stdsindy.}
	\label{tab:lorez_noisefree}
	\begin{center}
		\rowcolors{1}{}{lightgray}
		\begin{tabular}{|c|c|} \hline
			\bred{\large\rksindy} & \bred{\large\stdsindy} \\ \hline
			$	\begin{aligned}
				\dot{\tilde\bx}(t) &= -10.004 \tilde\bx(t) + 10.004 \tilde\by(t),\\
				\dot {\tilde\by}(t) &= 2.966\tilde\bx - 0.956 \tilde\by(t) -  7.953\tilde\bx(t)\tilde\bz(t) , \\
				\dot{\tilde\bz}(t) &= 7.944\tilde\bx(t)\tilde\by(t) - 2.669\tilde\bz(t) - 8.336
			\end{aligned}$ & $	\begin{aligned}
				\dot{\tilde\bx}(t) &= -9.983 \tilde\bx(t) + 9.983 \tilde\by(t),\\
				\dot {\tilde\by}(t) &= 2.912\tilde\bx - 0.922 \tilde\by(t) -  7.911\tilde\bx(t)\tilde\bz(t) , \\
				\dot{\tilde\bz}(t) &= 7.972\tilde\bx(t)\tilde\by(t) - 2.662\tilde\bz(t) - 8.313
			\end{aligned}$ \\ \hline
		\end{tabular}
	\end{center}
\end{table}
Next, we study the performance of the approaches under noisy measurements. For this, we add mean zero Gaussian noise of variance one. To employ \rksindy, we first apply a Savitzky-Golay filter \cite{savitzky1964smoothing} to denoise the time-history data, see \Cref{fig:lorez_denoise}. For \stdsindy~as well, we use the same filter to denoise the signal and approximate the derivative information. We plot the trajectories of the discovered models and ground-truth in \Cref{fig:lorez_noise} and observe that dynamics on an attractor is still intact; however, we note that the discovered equations are very different from the ground truth, see \Cref{tab:lorez_noise}. The learning can be improved by employing \Cref{algo:procedure2}, where we iteratively remove the smallest coefficient and determine the sparsest solutions by looking at the Pareto-front. However, it comes at a slightly higher computational cost. We discuss this approach more in detail in our following examples.

\begin{table}[htb]
	\begin{center}
		\rowcolors{1}{}{lightgray}
		\begin{tabular}{|c|c|} \hline
			\bred{\large\rksindy} & \bred{\large\stdsindy} \\ \hline
			$	\begin{aligned}
				\dot{\tilde\bx}(t) &= -9.016 \tilde\bx + 8.221 \tilde\by \\&~~ -1.675 \tilde\bx \tilde\bz + 0.895 \tilde\bx^3 \\&~~+ 0.603 \tilde\bx^2 \tilde\by + -0.579 \tilde\bx \tilde\by^2 \\
				\dot {\tilde\by}(t) &= 1.025 \tilde\by  -6.133 \tilde\bx \tilde\bz  -1.033 \tilde\by \tilde\bz \\
				\dot{\tilde\bz}(t) &= -8.345 1  -2.708 \tilde\bz + 7.971 \tilde\bx \tilde\by 
			\end{aligned}$ & $	\begin{aligned}
				\dot{\tilde\bx}(t) &= -8.842 \tilde\bx + 8.373 \tilde\by \\&~~ -3.107 \tilde\bx \tilde\bz + 1.065 \tilde\by \tilde\bz \\&~~+ 2.165 \tilde\bx^3 + -1.215 \tilde\bx^2 \tilde\by \\
				\dot {\tilde\by}(t) &= 1.811 \tilde\bx + -7.580 \tilde\bx \tilde\bz \\
				\dot{\tilde\bz}(t) &= -8.344 1 + -2.710 \tilde\bz + 7.950 \tilde\bx \tilde\by
			\end{aligned}$ \\ \hline
		\end{tabular}
	\end{center}
	\caption{Lorenz model: Discovered governing equations using \rksindy~and \stdsindy~from noisy measurements.}
	\label{tab:lorez_noise}
\end{table}

\subsection{Michaelis-Menten kinetics}\label{subsec:MM_model}
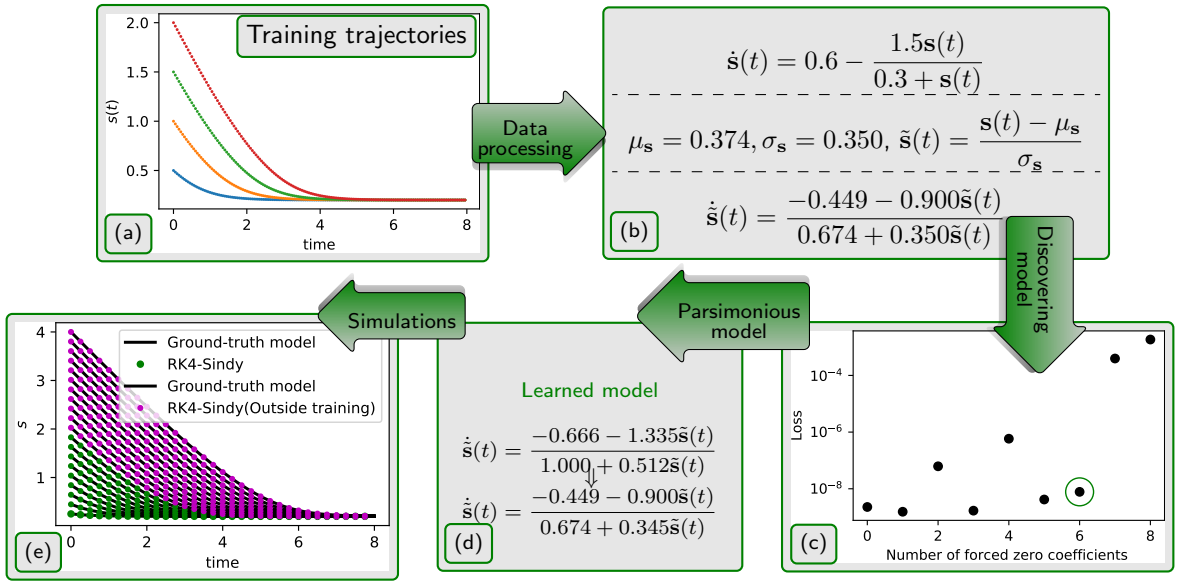
\begin{figure}[H]
	\begin{tikzpicture}[font=\sffamily]
		\node[ fill = white!90!black,draw = green!50!black, text = white, thick,rounded corners = 0.5ex, minimum height = 2cm] (IniCond) {	\includegraphics[width = 0.3\textwidth]{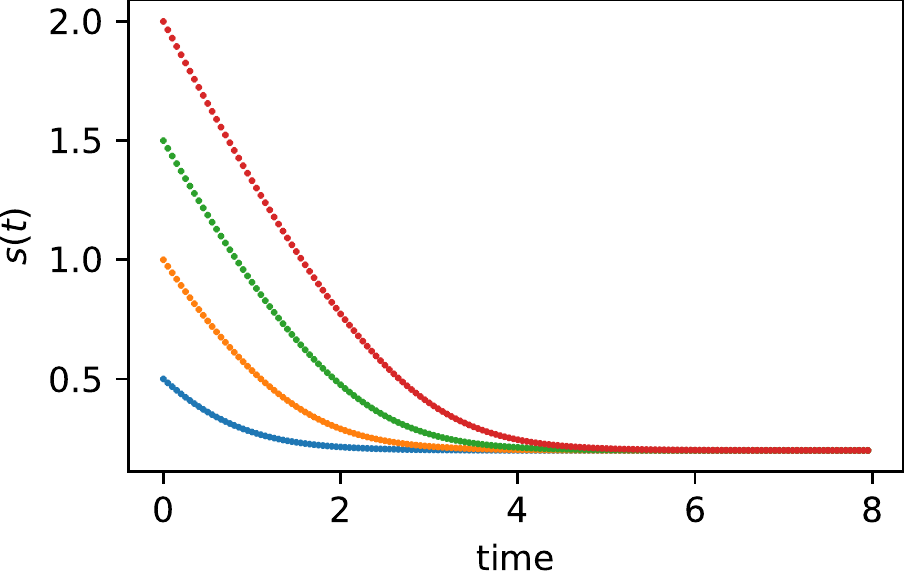}};
		\node[ fill = white!90!black,draw = green!50!black, above right = 1cm and -.75cm of IniCond.center, thick,rounded corners = 0.5ex] (data) {Training trajectories};
		\node[fill = white!90!black,draw = green!50!black, right = 1.5cm of IniCond, text = black, thick,rounded corners = 0.5ex] (MMeqns) {
			\begin{tabular}{c}
				$\begin{aligned}
					\dot\bs(t) = 0.6  - \dfrac{1.5 \bs(t)}{0.3 + \bs(t)}
				\end{aligned}
				$\\ \hdashline
				$\mu_\bs = 0.374,\sigma_\bs = 0.350$,
				$\tilde\bs(t) = \dfrac{\bs(t)-\mu_\bs}{\sigma_\bs}$\\ \hdashline
				$\begin{aligned}
					\dot{\tilde\bs}(t) = \dfrac{-0.449 - 0.900{\tilde\bs}(t)}{0.674 + 0.350\tilde\bs(t)}
				\end{aligned}
				$
			\end{tabular}
		};	
		\node[ fill = white!90!black,draw = green!50!black, below right = 2.5cm and -1cm of MMeqns.center, text = white, thick,rounded corners = 0.5ex] (pareto) {\includegraphics[width = 0.3\textwidth]{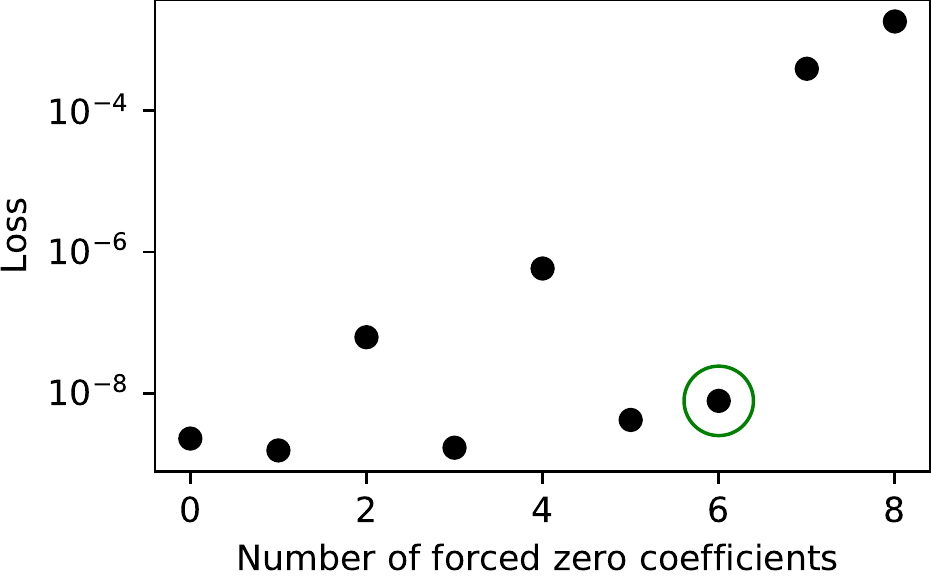}};	
		\node[ fill = white!90!black,draw = green!50!black, left = 0.5cm of pareto, thick,rounded corners = 0.5ex, minimum height = 3.3cm] (learnedmodel) {\footnotesize
			\begin{tabular}{c}
				{\color{green!50!black}Learned model}\\
				$\begin{aligned}
					\dot{\tilde\bs}(t) = \dfrac{-0.666 - 1.335{\tilde\bs}(t)}{1.000 + 0.512\tilde\bs(t)}
				\end{aligned}
				$ \\[-20pt] 
				\rotatebox{-90}{$\Rightarrow$}\\[-05pt]
				$\begin{aligned}
					\dot{\tilde\bs}(t) = \dfrac{-0.449 - 0.900{\tilde\bs}(t)}{0.674 + 0.345\tilde\bs(t)}
				\end{aligned}
				$
			\end{tabular}
		};	
		\node[ fill = white!90!black,draw = green!50!black, left = 0.5cm of learnedmodel, text = white, thick,rounded corners = 0.5ex] (simulations) {\includegraphics[width = 0.3\textwidth]{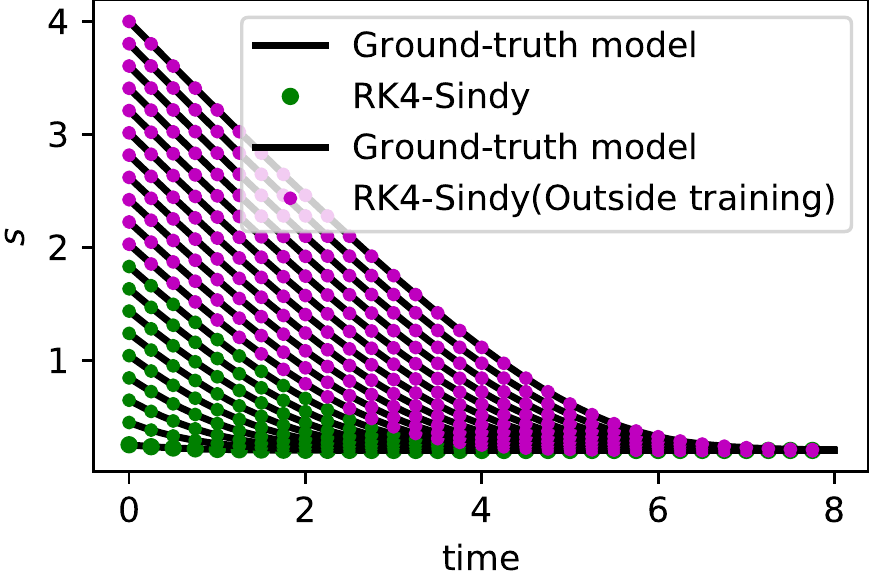}};	
		\node[ fill = none,draw = none, right = -0.1cm  of IniCond, thick,rounded corners = 0.5ex, minimum width = .1cm, minimum height = .1cm,inner sep=-0.15cm] () { \footnotesize
			\tikzfancyarrow[0.5cm]{\begin{tabular}{c} Data \\[-15pt] processing\end{tabular}}
		};
		\node[ fill = none,draw = none, below right = -0.5cm and -1.5cm of MMeqns, thick,rounded corners = 0.5ex, minimum width = .1cm, minimum height = .1cm,inner sep=-0.1cm] () {\footnotesize
			\rotatebox{-90}{\tikzfancyarrow[0.5cm]{\begin{tabular}{c} Discovering  \\[-15pt] model\end{tabular}}}
		};
		\node[ fill = none,draw = none, above left = -0.5cm and -0.3cm  of pareto, thick,rounded corners = 0.5ex, minimum width = .1cm, minimum height = .1cm,inner sep=-0.1cm] () {\footnotesize
			\rotatebox{180}{\tikzfancyarrow[0.5cm]{ \rotatebox{180}{\begin{tabular}{c} Parsimonious   \\[-15pt] model\end{tabular}}}}
		};
		\node[ fill = none,draw = none, above left = -0.35cm and -0.3cm  of learnedmodel, thick,rounded corners = 0.5ex, minimum width = .1cm, minimum height = .1cm,inner sep=-0.1cm] () {\footnotesize
			\rotatebox{180}{\tikzfancyarrow[0.5cm]{ \rotatebox{180}{\begin{tabular}{c} Simulations\end{tabular}}}}
		};
		\node[fill = white!90!black,draw = green!50!black, below left = -0.65cm and -0.75cm  of IniCond, thick,rounded corners = 0.5ex] () {\footnotesize (a)
		};
		\node[fill = white!90!black,draw = green!50!black, below left = -0.65cm and -0.75cm  of MMeqns, thick,rounded corners = 0.5ex] () {\footnotesize (b)
		};
		\node[fill = white!90!black,draw = green!50!black, below left = -0.65cm and -0.75cm  of pareto, thick,rounded corners = 0.5ex] () {\footnotesize (c)
		};
		\node[fill = white!90!black,draw = green!50!black, below left = -0.65cm and -0.75cm  of learnedmodel, thick,rounded corners = 0.5ex] () {\footnotesize (d)
		};
		\node[fill = white!90!black,draw = green!50!black, below left = -0.65cm and -0.75cm  of simulations, thick,rounded corners = 0.5ex] () {\footnotesize (e)
		};
	\end{tikzpicture}
	\caption{Michaelis-Menten kinetics: In the first step, we have collected data for $4$ initial conditions at a time-stepping $\dt=5\cdot10^{-2}$. In the second step, we performed data-processing to normalize the data using the mean and standard deviation. In the third step, we employed \rksindy~(\Cref{algo:procedure2}) to discover the most parsimonious model. For this, we observe the Pareto front and pick the model that has the best fit to the data yet having the maximum number of zero coefficients. We then compare the discovered model with the ground truth and find that the proposed methodology could find precise candidates from the polynomial dictionary. The corresponding coefficients have less than $1\%$ errors.}
	\label{fig:MM_model}
\end{figure}

To illustrate \rksindy~to discover governing equations that are given by a ratio of two nonlinear functions, we consider arguably the most well-known model for an enzyme kinetics, namely Michaelis-Menten model \cite{michaelis1913kinetik,johnson2011original}. The model explains the dynamics of binding and unbinding of enzyme with an substrate $\bs$. In a simplistic way, the dynamics are governed by \cite{briggs1925further}:
\begin{equation}
	\dot\bs(t) = 0.6  - \dfrac{1.5 \bs(t)}{0.3 + \bs(t)}.
\end{equation}
As a first step, we generate data using four initial conditions  $\{0.5,1.0,1.5,2.0\}$. We collect data at a time-step $\dt = 5\cdot10^{-2}$, see \Cref{fig:MM_model}(a). Typically, governing equations, explaining biological process have rational functions. Therefore, we aim at discovering the enzyme kinetics model by assuming a rational form as shown in \eqref{eq:ratio_fun}, i.e.,  the gradient field of $\bs(t)$ takes the form $\tfrac{\bg(\bs(t))}{1+ \bh(\bs(t))}$. 

Next, in order to identify $\bg(\bs)$ and $\bh(\bs)$, we construct the polynomial dictionaries, containing terms up to degrees  $4$. After that, we employ \rksindy~to identify the precise features from the dictionaries to characterize  $\bg$ and $\bh$. Moreover, we apply the iterative thresholding approach discussed in \Cref{algo:procedure2}, in contrast to previously considered examples where a fixed thresholding is applied. Note that the success of \rksindy~approach not only depends on a dictionary containing candidate features but the quality of data. We have marked that the dictionary data matrix's conditioning improves when data are normalized to mean-zero and variance-one. It is crucial for polynomial basis in the dictionary. For this example, we normalize the data before employing \rksindy. It means that the transformation is done as follows:
\begin{equation}
	\tilde\bs(t) = \dfrac{\tilde{\bs} - \mu_\bs}{{\sigma_\bs}},
\end{equation}
where $\mu_\bs$ and  $\sigma_\bs$ are the mean and standard deviation of the collected data. Next, using the normalized data, we learn the governing equation, describing the dynamics of $\tilde\bs(t)$. Since we consider dictionaries for $\bg$ and $\bh$, containing polynomials of degree $4$, there are total $9$ coefficients. To identify the correct model while employing \Cref{algo:procedure2}, we keep track of the loss (data-fidelity) and the number of non-zero coefficients, which is shown in \Cref{fig:MM_model}(c). This allows us to built a Pareto front for the optimization problem and choose the most parsimonious model that describes the dynamics present in collected data. One of the most attractive features of learning parsimonious models is to avoid over-fitting and generalizing better in regions in which data are not collected. It is precisely what we observed as well. As shown in  \Cref{fig:MM_model}(e), the learned model predicts dynamics very accurately in the region far away from the training one.

Next, we study the performance of the method under noisy measurements. For this, we corrupt the collected data using zero-mean Gaussian noise of variance $\sigma = 2\cdot 10^{-2}$. Then, we process the data by first employing a noise-reduction filter, namely Savitzky-Golay, followed by normalizing the data. In the third step, we focus on learning the most parsimonious model by picking appropriate candidates from the polynomial dictionary. Remarkably, the method allows us to find a model with correct features from the dictionary and coefficient accuracy up to $5\%$. Furthermore, the model faithfully generalizes to regimes outside the training, even using noisy measurements. 

\begin{SCfigure}[\sidecaptionrelwidth][h!]
	\begin{tikzpicture}[font=\sffamily]
		\node[ fill = white!90!black,draw = green!50!black, text = white, thick,rounded corners = 0.5ex] (IniCond) {	\includegraphics[height = 3.0cm, width = 0.3\textwidth]{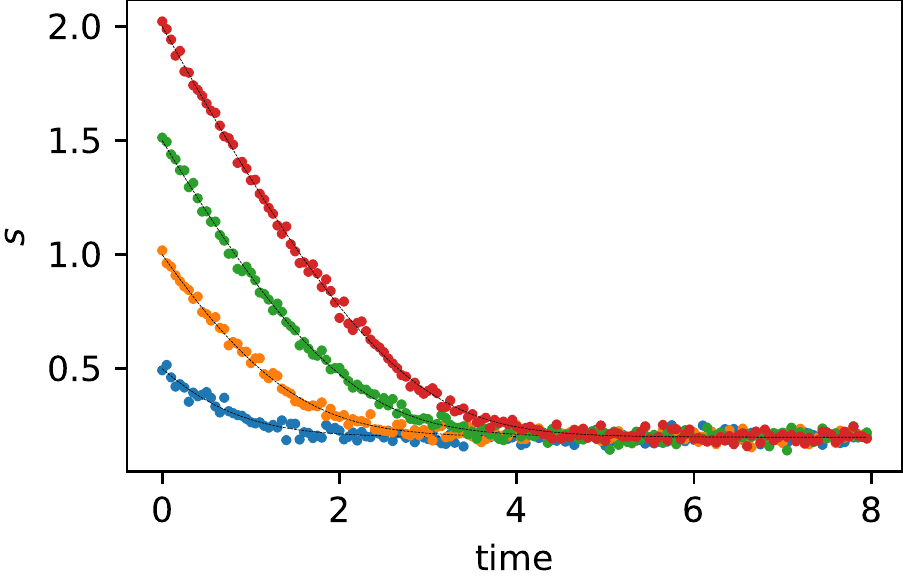}};
		\node[ fill = white!90!black,draw = green!50!black, above right = 1cm and -.5cm of IniCond.center, thick,rounded corners = 0.5ex] (data) {Noisy measurements};
		\node[fill = white!90!black,draw = green!50!black, right = 1.0cm of IniCond, text = black, thick,rounded corners = 0.5ex] (MMeqns) {
			\includegraphics[height = 3.0cm, width = 0.3\textwidth]{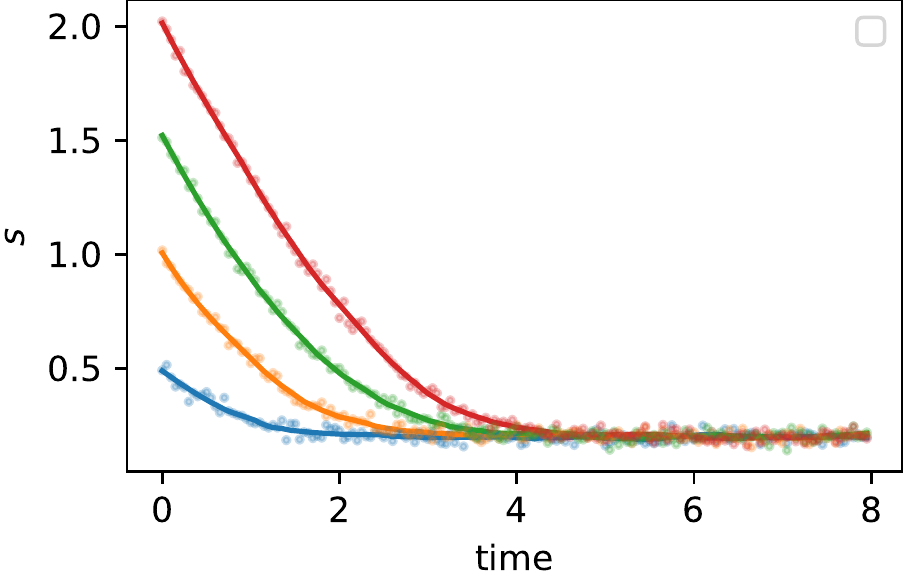}};
		\node[ fill = white!90!black,draw = green!50!black, below  = 0.2cm of MMeqns, text = white, thick,rounded corners = 0.5ex] (pareto) {\includegraphics[height = 3.0cm, width = 0.3\textwidth]{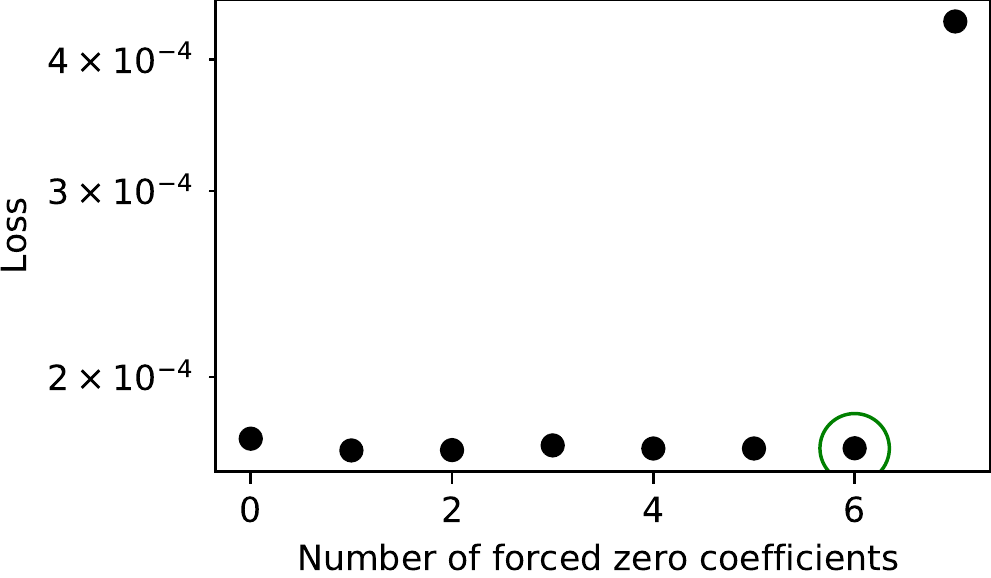}};	
		\node[ fill = white!90!black,draw = green!50!black, left = 1.0cm of pareto, text = white, thick,rounded corners = 0.5ex] (simulations) {\includegraphics[height = 3.0cm, width = 0.3\textwidth]{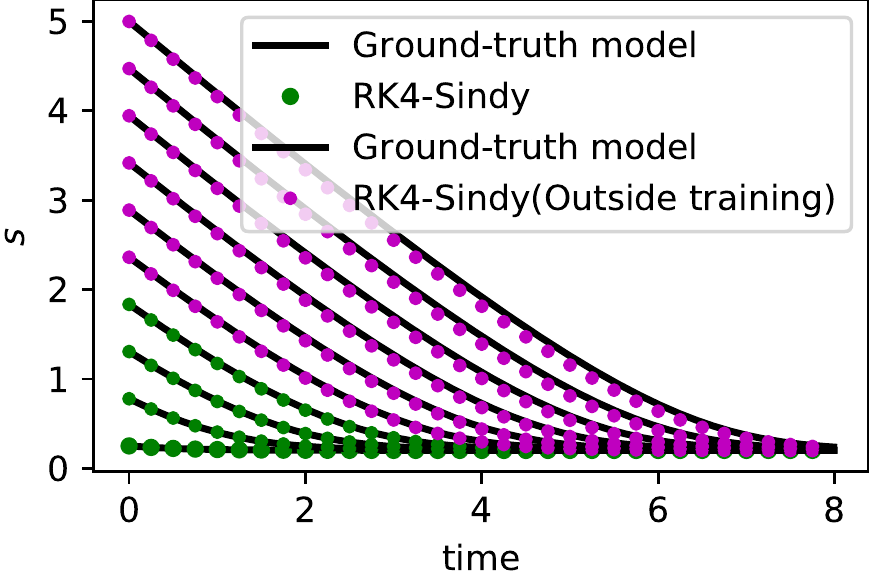}};	
		\node[ fill = none,draw = none, right = -0.5cm  of IniCond, thick,rounded corners = 0.5ex, minimum width = .1cm, minimum height = .1cm,inner sep=-0.15cm] () { \footnotesize
			\tikzfancyarrow[0.5cm]{\begin{tabular}{c} Data \\[-15pt] processing\end{tabular}}
		};
		\node[ fill = none,draw = none, below right = -0.1cm and -1.5cm of MMeqns, thick,rounded corners = 0.5ex, minimum width = .1cm, minimum height = .1cm,inner sep=-0.1cm] () {\footnotesize
			\rotatebox{-90}{\tikzfancyarrow[0.5cm]{\begin{tabular}{c} Discovering  \\[-15pt] model\end{tabular}}}
		};
		\node[ fill = none,draw = none, above left = -2.7cm and -0.2cm  of pareto, thick,rounded corners = 0.5ex, minimum width = .1cm, minimum height = .1cm,inner sep=-0.1cm] () {\footnotesize
			\rotatebox{180}{\tikzfancyarrow[0.5cm]{ \rotatebox{180}{\begin{tabular}{c} Parsimonious   \\[-15pt] model\end{tabular}}}}
		};
	\end{tikzpicture}
	\caption{Michaelis-Menten kinetics:  The figure demonstrates the necessary steps to uncover the most parsimonious model using noisy measurement data. It also testifies to the impressive capability of discovering the most parsimonious model -- that is, they even generalize beyond the training regime.}
\end{SCfigure}
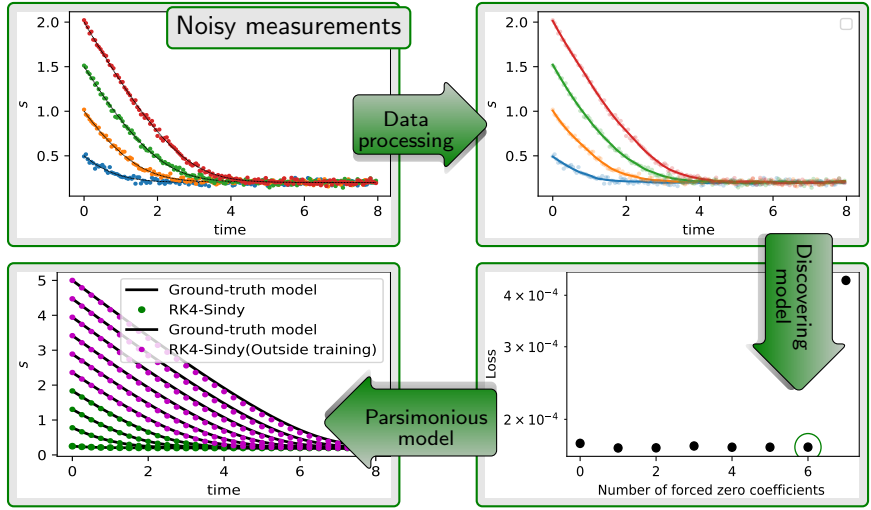

\subsection{Hopf normal form}\label{subsec:hopf}
\begin{SCfigure}[\sidecaptionrelwidth][!tb]
	\captionsetup{format=plain}
	\begin{tikzpicture}[font=\sffamily]
		\node[ fill = white!100!black,draw = green!50!black, text = white, thick,rounded corners = 0.5ex] (testmodel1) {	
			\includegraphics[height = 3.75cm, width = 0.34\textwidth]{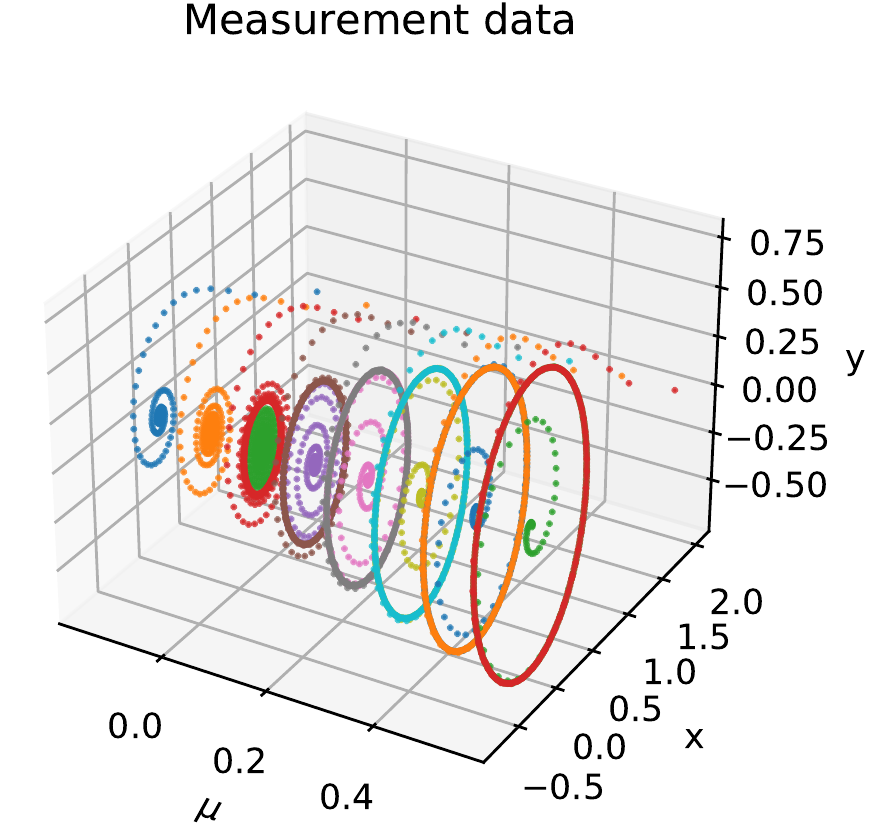}
			\includegraphics[height = 3.75cm, width = 0.34\textwidth]{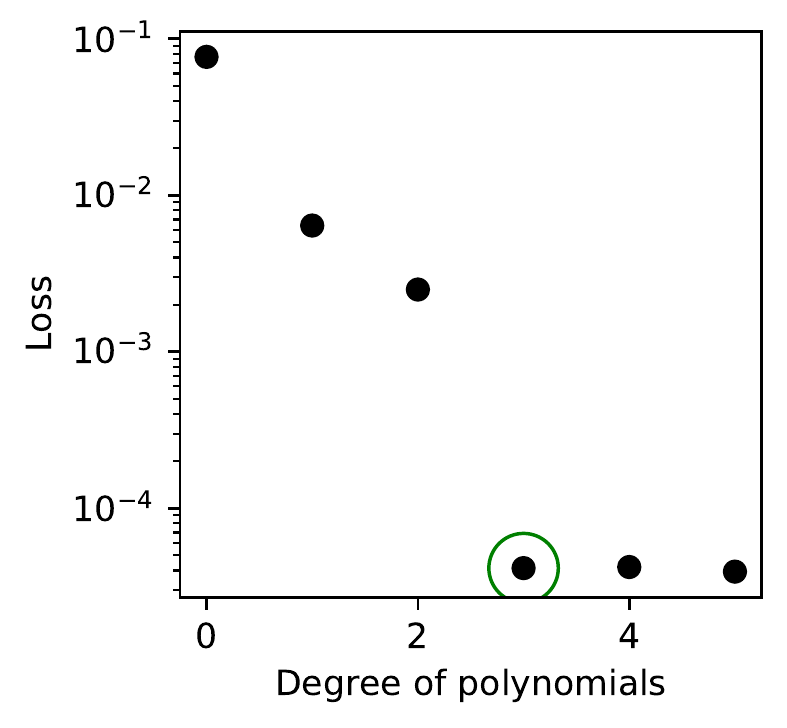}
		};
		\node[ fill = white!100!black,draw = green!50!black, text = white, thick,below = 0cm of testmodel1, rounded corners = 0.5ex] (testmodel2) {	
			\includegraphics[height = 3.75cm, width = 0.34\textwidth]{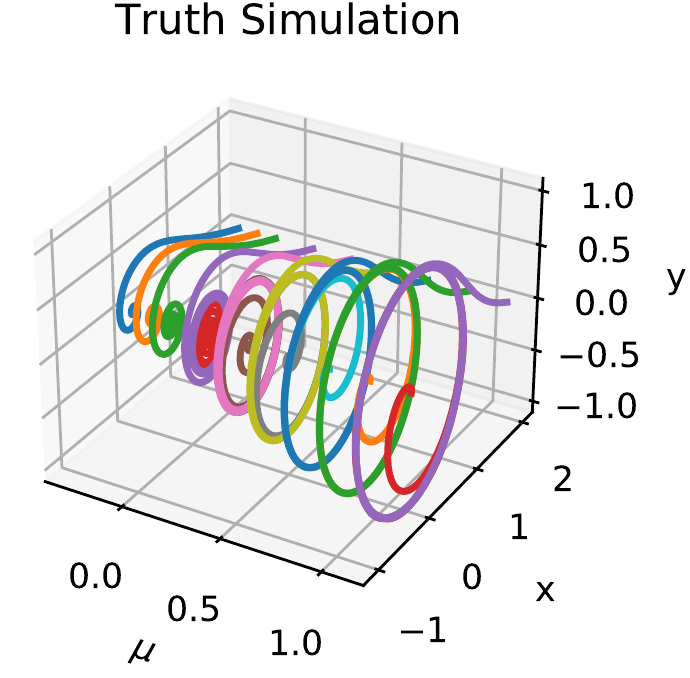}
			\includegraphics[height = 3.75cm, width = 0.34\textwidth]{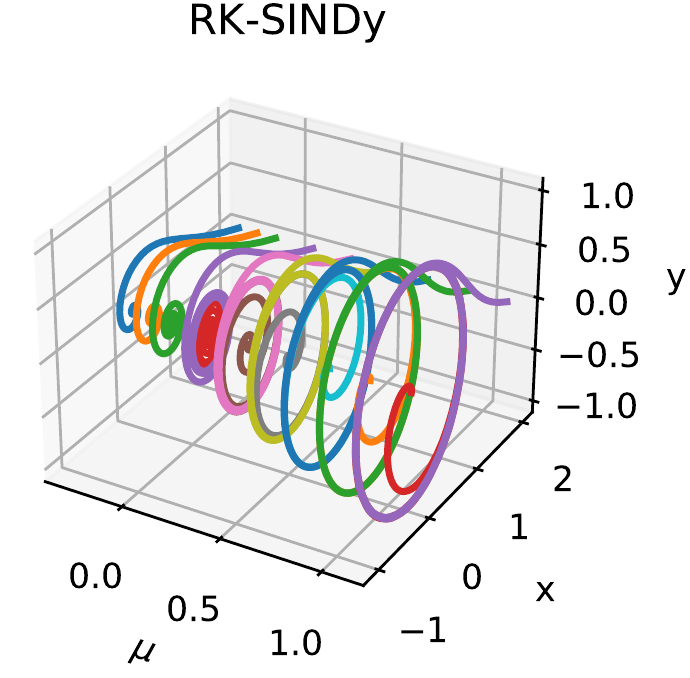}};
	\end{tikzpicture}
	\caption{Hopf normal form: The top-left figure shows the noisy measurements that are obtained using various parameter $\mu$. To identify correct degree polynomial basis in the dictionary, we do a assessment test, indicating degree-3 polynomial are sufficient to describe dynamics (top-right). The bottom figures shows a comparison of simulations of the ground truth model and  identified models for the parameter $\mu$, illustrating  the capability of generalizing beyond the training parameter regime.}
	\label{fig:hopf}
\end{SCfigure}

In our last example, we study discovering parameterized differential equations from noisy measurements. Many real-world dynamical processes have system parameters, and depending on them, the system may exhibit very distinctive dynamical behaviors. To illustrate the efficiency of \rksindy~to discover parametric equations, we consider the Hopf system
\begin{equation}
	\begin{aligned}
		\dot\bx(t) &= \mu \bx(t) - \omega \by(t) - \bA\bx(t)\left(\bx(t)^2 + \by(t)^2\right),\\
		\dot\by(t) &= \omega \bx(t) + \mu \by(t) - \bA\by(t)\left(\bx(t)^2 + \by(t)^2\right)
	\end{aligned}
\end{equation}
that exhibits bifurcation with respect to the parameter $\mu$. For this example, we collect measurements for eight different parameter values $\mu$ at a time-step $0.2$ by fixing $\omega = 1$ and $\bA = 1$. Then, we corrupt the measurement data by adding a Gaussian sensor noise that is shown in \Cref{fig:hopf} (left top).  Next, we aim at constructing a symbolic polynomial dictionary $\Phi$ by including the parameter $\mu$ as the dependent variables. While building a polynomial dictionary, it is important to choose the degree of the polynomial as well. Moreover, it is known that the polynomial basis becomes numerically unstable as the degree increases. Hence, solving optimization problem that discovers governing equations becomes challenging. With mean of this example, we discuss an assessment test to choose the appropriate degree of the polynomial in the dictionary. Essentially, we inspect  data fidelity with respect to the degree of the polynomial in the dictionary.  When the dictionary contains all essential polynomial features, then a sharp drop in the error is expected. 
We observe in  \Cref{fig:hopf} (right-top) a sharp drop in the error at the degree $3$, and the error remains almost the same even when higher polynomial features are added. It indicates that polynomial degree $3$ is sufficient to describe the dynamics. 
Thereafter, using the dictionary containing degree $3$ polynomial features, we seek to identify the minimum number of features from the dictionary that explains the underlying dynamics. We achieve this by employing \rksindy, and compare the performance with \stdsindy. We note down the discovered governing equations in \Cref{tab:hopf}, where we notice an impressive performance of \rksindy~to discover the exact form of the underlying parametric equations, and the coefficients are up to $1\%$ accurate. On the other hand, \stdsindy~is not able to identify the correct form of the model.  Furthermore, we compare the discovered model simulations using \rksindy~with ground truth beyond the training regime of the parameter $\mu$ in \Cref{fig:hopf} (bottom). It exposes the strength of the parsimonious and interpretable discovered models.

\begin{table}[tb]
	\begin{tabular}{|c|c|} \hline
		Method & Discovered model \\ \hline
		\bred{\rksindy}&  	$\begin{aligned}
			\dot\bx(t) &= 1.001\mu \bx(t) - 1.001 \by(t) - 0.996\bx(t) \left(\bx(t)^2 + \by(t)^2\right)\\
			\dot\by(t) &= 1.001 \bx(t) + 1.010\mu \by(t) - 1.006 \bx(t)^2\by(t)  -1.004  \by(t)^3
		\end{aligned}$ \\ \hline
		\bred{\stdsindy}& 	{\scriptsize $\begin{aligned}
				\dot\bx(t) &= -0.961 \by(t) + 0.719 \mu \bx(t) + 0.822 \mu\by(t) -0.735 \bx(t)^3  -1.044 \bx(t)^2 \by -0.686 \bx(t) \by(t)^2  -0.846 \by(t)^3 \\ 
				\dot\by(t) &= 0.986 \bx(t) + 0.899 \mu\by(t) -0.882 \bx(t)^2 \by(t) -0.904 \by(t)^3.
			\end{aligned}$}\\ \hline
	\end{tabular}
	\caption{Hopf normal form: Here, we report discovered governing equations using noisy measurement data, representing dynamics of Hopf bifurcation and notice that \rksindy~recovers the Hopf normal form very accurately; on the other hand, \stdsindy~breaks down.}
	\label{tab:hopf}
\end{table}

\section{Discussion}
This work has manifested a compelling approach (\rksindy) to discover nonlinear differential equations without imposing any prior structure on models. For this, we have blended sparsity-promoting identification with a numerical integration scheme, namely, Runge-Kutta  $4^{\text{th}}$-order scheme. The beauty of the proposed methodology is that we do not require derivative information at any stage, notwithstanding we still discover differential equations. Hence, the proposed algorithm differs from previously suggested sparsity-promoting identification methods in the literature in this aspect.  Consequently, we expect \rksindy~to perform better under sparsely sampled and corrupted data. We have demonstrated the efficiency of the approach on a variety of examples, namely linear and nonlinear damped oscillators, a model describing neural dynamics, chaotic behavior, and parametric differential equations. We have accurately discovered the Fitz-Hugh Nagumo model that describes the activation and de-activation of neurons.  We have also illustrated the identification of the Lorenz model and have shown that dynamics of identified models are intact on an attractor as it is more important for chaotic dynamics. The example of Michaelis-Menten Kinetics highlights that the proposed algorithm can discover models that are given by a ratio of two functions. The example also shows the power of determining parsimonious models -- that is, their generalization beyond the region in which data are selected. Furthermore, we have demonstrated the remarkable robustness of the proposed \rksindy~algorithm to sparsely-sampled data and to corrupted measurement data. In the case of large noise, a noise-reduction filter such as Savitzky-Golay helps to improve the quality of discovered governing equations. We have also reported a comparison with the sparse identification approach \cite{brunton2016sparse} and have observed the out-performance of \rksindy~over the latter approach. 

This work opens many exciting doors for further research from both theory and practical perspectives. Since the approach aims at selecting the correct features from a dictionary containing a high-dimensional nonlinear feature basis, the construction of these feature bases in a dictionary plays a significant role in determining the success of the approach.  There is no straightforward answer to this obstacle; however, there is some expectation that meaningful features may be constructed with the help of experts and empirical knowledge, or at least they may be realized in raw forms by them.  Furthermore, we have solved the optimization problem \eqref{eq:opt_problem} using a gradient-based method. We have observed that if feature functions in the dictionary are similar for given data, then the convergence is slow, and sometimes it even fails and is stuck in a local minimum. In other words, the incoherency between the feature functions is low. Hence, there is a need for the normalization step. In Subsections \ref{subsec:Lorenz} and \ref{subsec:MM_model}, we have employed a normalization step to improve incoherency. However, it is worth investigating better-suited strategies to normalize either data or feature spaces as a pre-processing step so that sparsity in the feature space remains intact.    In addition to these, a thorough study on the performance of various noise-reduction methods would provide deep insights into their appropriateness to \rksindy, despite we noticed a good performance of the Sabitzky-Golay filter to reduced noise in our results. 

Methods discovering interpretable models that generalize well beyond the training regime are limited, and the proposed method \rksindy~is among these. Additionally, approaches discovering governing equations that also obey physical laws are even rarer.  A very recent paper \cite{willcox2021imperative} has stressed that discovering/learning models can be made even more efficient by incorporating the laws of nature in the course of discovering equations.  A solid example comes from the discovering biological networks that often follow the mass-conversation law. Therefore, integrating physical laws in discovering models and sparse identification will hopefully shape the future of discovering explainable and generalizable differential equations. 

%
%
\addcontentsline{toc}{section}{References}
\bibliographystyle{plain}
\bibliography{mybib.bib}

\begin{thebibliography}{10}

\bibitem{ascher1998computer}
Uri~M Ascher and Linda~R Petzold.
\newblock {\em Computer methods for ordinary differential equations and
  differential-algebraic equations}, volume~61.
\newblock SIAM, 1998.

\bibitem{beck2013sparsity}
Amir Beck and Yonina~C Eldar.
\newblock Sparsity constrained nonlinear optimization: Optimality conditions
  and algorithms.
\newblock {\em {SIAM} J. Optim.}, 23(3):1480--1509, 2013.

\bibitem{bongard2007automated}
Josh Bongard and Hod Lipson.
\newblock Automated reverse engineering of nonlinear dynamical systems.
\newblock {\em Proc. Nat. Acad. Sci. U.S.A.}, 104(24):9943--9948, 2007.

\bibitem{briggs1925further}
George~Edward Briggs.
\newblock A further note on the kinetics of enzyme action.
\newblock {\em Biochem. J.}, 19(6):1037, 1925.

\bibitem{brunton2016discovering}
Steven~L Brunton, Joshua~L Proctor, and J~Nathan Kutz.
\newblock Discovering governing equations from data by sparse identification of
  nonlinear dynamical systems.
\newblock {\em Proc. Nat. Acad. Sci. U.S.A.}, 113(15):3932--3937, 2016.

\bibitem{brunton2016sparse}
Steven~L Brunton, Joshua~L Proctor, and J~Nathan Kutz.
\newblock Sparse identification of nonlinear dynamics with control ({SINDYc}).
\newblock {\em IFAC-PapersOnLine}, 49(18):710--715, 2016.

\bibitem{candes2006robust}
Emmanuel~J Cand{\`e}s, Justin Romberg, and Terence Tao.
\newblock Robust uncertainty principles: {E}xact signal reconstruction from
  highly incomplete frequency information.
\newblock {\em IEEE Trans. Inform. Theory}, 52(2):489--509, 2006.

\bibitem{candes2006stable}
Emmanuel~J Candès, Justin~K Romberg, and Terence Tao.
\newblock Stable signal recovery from incomplete and inaccurate measurements.
\newblock {\em Comm. Pure Appl. Math.}, 59(8):1207--1223, 2006.

\bibitem{champion2020unified}
Kathleen Champion, Peng Zheng, Aleksandr~Y Aravkin, Steven~L Brunton, and
  J~Nathan Kutz.
\newblock A unified sparse optimization framework to learn parsimonious
  physics-informed models from data.
\newblock {\em IEEE Access}, 8:169259--169271, 2020.

\bibitem{chartrand2011numerical}
Rick Chartrand.
\newblock Numerical differentiation of noisy, nonsmooth data.
\newblock {\em Intern. Scholarly Res. Notices}, 2011, 2011.

\bibitem{crutchfield1987equations}
James~P Crutchfield and Bruce~S McNamara.
\newblock Equations of motion from a data series.
\newblock {\em Complex Sys.}, 1(417-452):121, 1987.

\bibitem{daniels2015efficient}
Bryan~C Daniels and Ilya Nemenman.
\newblock Efficient inference of parsimonious phenomenological models of
  cellular dynamics using {S}-systems and alternating regression.
\newblock {\em {PLoS One}}, 10(3):e0119821, 2015.

\bibitem{desilva2020}
Brian de~Silva, Kathleen Champion, Markus Quade, Jean-Christophe Loiseau,
  J.~Kutz, and Steven Brunton.
\newblock {PySINDy}: A {P}ython package for the sparse identification of
  nonlinear dynamical systems from data.
\newblock {\em J. Open Source Software}, 5(49):2104, 2020.

\bibitem{donoho2006compressed}
David~L Donoho.
\newblock Compressed sensing.
\newblock {\em IEEE Trans. Inform. Theory}, 52(4):1289--1306, 2006.

\bibitem{fitzhugh1955mathematical}
Richard FitzHugh.
\newblock Mathematical models of threshold phenomena in the nerve membrane.
\newblock {\em The Bulletin Math. Biophys.}, 17(4):257--278, 1955.

\bibitem{friedman2001elements}
Jerome Friedman, Trevor Hastie, Robert Tibshirani, et~al.
\newblock {\em The elements of statistical learning}, volume~1.
\newblock Springer, 2001.

\bibitem{gavish2017optimal}
Matan Gavish and David~L Donoho.
\newblock Optimal shrinkage of singular values.
\newblock {\em IEEE Trans. Inform. Theory}, 63(4):2137--2152, 2017.

\bibitem{he2016deep}
Kaiming He, Xiangyu Zhang, Shaoqing Ren, and Jian Sun.
\newblock Deep residual learning for image recognition.
\newblock In {\em Proc. IEEE Conf. Comp. Vision Patt. Recog.}, pages 770--778,
  2016.

\bibitem{huang2017densely}
Gao Huang, Zhuang Liu, Laurens Van Der~Maaten, and Kilian~Q Weinberger.
\newblock Densely connected convolutional networks.
\newblock In {\em Proc. IEEE Conf. Comp. Vision Patt. Recog.}, pages
  4700--4708, 2017.

\bibitem{james2013introduction}
Gareth James, Daniela Witten, Trevor Hastie, and Robert Tibshirani.
\newblock {\em An introduction to statistical learning}, volume 112.
\newblock Springer, 2013.

\bibitem{johnson2011original}
Kenneth~A Johnson and Roger~S Goody.
\newblock The original {M}ichaelis constant: translation of the 1913
  {Michaelis--Menten} paper.
\newblock {\em Biochemistry}, 50(39):8264--8269, 2011.

\bibitem{jordan2015machine}
Michael~I Jordan and Tom~M Mitchell.
\newblock Machine learning: {T}rends, perspectives, and prospects.
\newblock {\em Science}, 349(6245):255--260, 2015.

\bibitem{kantz2004nonlinear}
Holger Kantz and Thomas Schreiber.
\newblock {\em Nonlinear Time Series Analysis}, volume~7.
\newblock Cambridge University Press, 2004.

\bibitem{kevrekidis2003equation}
Ioannis~G Kevrekidis, C~William Gear, James~M Hyman, Panagiotis~G Kevrekidis,
  Olof Runborg, Constantinos Theodoropoulos, et~al.
\newblock Equation-free, coarse-grained multiscale computation: {E}nabling
  mocroscopic simulators to perform system-level analysis.
\newblock {\em Comm. Math. Sci.}, 1(4):715--762, 2003.

\bibitem{NarP90}
S~Narendra Kumpati and Parthasarathy Kannan.
\newblock Identification and control of dynamical systems using neural
  networks.
\newblock {\em IEEE Trans. Neural Networks}, 1(1):4--27, 1990.

\bibitem{lennart1999system}
Lennart Ljung.
\newblock {\em System Identification: Theory for the User}.
\newblock Prentice Hall, NJ, 1999.

\bibitem{lorenz1963deterministic}
Edward~N Lorenz.
\newblock Deterministic nonperiodic flow.
\newblock {\em J. Atmospheric Sci.}, 20(2):130--141, 1963.

\bibitem{mangan2016inferring}
Niall~M Mangan, Steven~L Brunton, Joshua~L Proctor, and J~Nathan Kutz.
\newblock Inferring biological networks by sparse identification of nonlinear
  dynamics.
\newblock {\em IEEE Trans. Molecular, Biological and Multi-Scale Comm.},
  2(1):52--63, 2016.

\bibitem{marx2013big}
Vivien Marx.
\newblock The big challenges of big data.
\newblock {\em Nature}, 498(7453):255--260, 2013.

\bibitem{michaelis1913kinetik}
Leonor Michaelis and Maud~L Menten.
\newblock Die kinetik der invertinwirkung.
\newblock {\em Biochem. z}, 49(333-369):352, 1913.

\bibitem{ozolicnvs2013compressed}
Vidvuds Ozoli{\c{n}}{\v{s}}, Rongjie Lai, Russel Caflisch, and Stanley Osher.
\newblock Compressed modes for variational problems in mathematics and physics.
\newblock {\em Proc. Nat. Acad. Sci. U.S.A.}, 110(46):18368--18373, 2013.

\bibitem{proctor2014exploiting}
Joshua~L Proctor, Steven~L Brunton, Bingni~W Brunton, and JN~Kutz.
\newblock Exploiting sparsity and equation-free architectures in complex
  systems.
\newblock {\em Europ. Phy. J. Spec. Top.}, 223(13):2665--2684, 2014.

\bibitem{raissi2019physics}
Maziar Raissi, Paris Perdikaris, and George~E Karniadakis.
\newblock Physics-informed neural networks: A deep learning framework for
  solving forward and inverse problems involving nonlinear partial differential
  equations.
\newblock {\em J. Comput. Phys.}, 378:686--707, 2019.

\bibitem{raissi2018multistep}
Maziar Raissi, Paris Perdikaris, and George~Em Karniadakis.
\newblock Multistep neural networks for data-driven discovery of nonlinear
  dynamical systems.
\newblock {\em arXiv preprint arXiv:1801.01236}, 2018.

\bibitem{rudy2019deep}
Samuel~H Rudy, J~Nathan Kutz, and Steven~L Brunton.
\newblock Deep learning of dynamics and signal-noise decomposition with
  time-stepping constraints.
\newblock {\em J. Comput. Phys.}, 396:483--506, 2019.

\bibitem{savitzky1964smoothing}
Abraham Savitzky and Marcel~JE Golay.
\newblock Smoothing and differentiation of data by simplified least squares
  procedures.
\newblock {\em Analytical Chem.}, 36(8):1627--1639, 1964.

\bibitem{schmidt2009distilling}
Michael Schmidt and Hod Lipson.
\newblock Distilling free-form natural laws from experimental data.
\newblock {\em science}, 324(5923):81--85, 2009.

\bibitem{schmidt2011automated}
Michael~D Schmidt, Ravishankar~R Vallabhajosyula, Jerry~W Jenkins, Jonathan~E
  Hood, Abhishek~S Soni, John~P Wikswo, and Hod Lipson.
\newblock Automated refinement and inference of analytical models for metabolic
  networks.
\newblock {\em Phy. Biology}, 8(5):055011, 2011.

\bibitem{SuyVdM96}
Johan~AK Suykens, Joos~PL Vandewalle, and Bart~L de~Moor.
\newblock {\em Artificial Neural Networks for Modelling and Control of
  Non-Linear Systems}.
\newblock Springer, 1996.

\bibitem{tibshirani1996regression}
Robert Tibshirani.
\newblock Regression shrinkage and selection via the lasso.
\newblock {\em J. Royal Stat. Soc.: Series B (Methodological)}, 58(1):267--288,
  1996.

\bibitem{tropp2007signal}
Joel~A Tropp and Anna~C Gilbert.
\newblock Signal recovery from random measurements via orthogonal matching
  pursuit.
\newblock {\em IEEE Trans. Inform. Theory}, 53(12):4655--4666, 2007.

\bibitem{VanM96}
Peter {Van~Overschee} and Bart {de Moor}.
\newblock {\em Subspace Identification of Linear Systems: Theory,
  Implementation, Applications}.
\newblock Kluwer Academic Publishers, 1996.

\bibitem{wang2011predicting}
Wen-Xu Wang, Rui Yang, Ying-Cheng Lai, Vassilios Kovanis, and Celso Grebogi.
\newblock Predicting catastrophes in nonlinear dynamical systems by compressive
  sensing.
\newblock {\em Phy. Rev. Letters}, 106(15):154101, 2011.

\bibitem{willcox2021imperative}
Karen~E Willcox, Omar Ghattas, and Patrick Heimbach.
\newblock The imperative of physics-based modeling and inverse theory in
  computational science.
\newblock {\em Nature Comput. Sci.}, 1(3):166--168, 2021.

\bibitem{yang2016sparse}
Zhuoran Yang, Zhaoran Wang, Han Liu, Yonina Eldar, and Tong Zhang.
\newblock Sparse nonlinear regression: Parameter estimation under nonconvexity.
\newblock In {\em Intern. Conf. on Machine Learning}, pages 2472--2481. PMLR,
  2016.

\bibitem{ye2015equation}
Hao Ye, Richard~J Beamish, Sarah~M Glaser, Sue~CH Grant, Chih-hao Hsieh,
  Laura~J Richards, Jon~T Schnute, and George Sugihara.
\newblock Equation-free mechanistic ecosystem forecasting using empirical
  dynamic modeling.
\newblock {\em Proc. Nat. Acad. Sci. U.S.A.}, 112(13):E1569--E1576, 2015.

\end{thebibliography}

\end{document}